\documentclass{article} 
\usepackage{arxiv,times}

\usepackage{amsmath,amsfonts,bm}
\usepackage{graphicx}
\usepackage{booktabs, multirow}
\usepackage[table]{xcolor}
\usepackage{caption}
\usepackage{xcolor}

\def\eqref#1{equation~\ref{#1}}

\def\1{\bm{1}}

\DeclareMathAlphabet{\mathsfit}{\encodingdefault}{\sfdefault}{m}{sl}
\SetMathAlphabet{\mathsfit}{bold}{\encodingdefault}{\sfdefault}{bx}{n}

\usepackage{hyperref}
\usepackage{url}

\title{A Diagnostic Study of World-Space Human Recovery Under High Camera Dynamics}

\author{Hao-Jen Chien \\
  National Yang Ming Chiao Tung University \\
  \texttt{hjchien90190@nycu.edu.tw} \\
  \And
  Wei-Lun Chao \\
  Boston University \\
  \texttt{chao209@bu.edu} \\
  \And
  Yu-Lun Liu \\
  National Yang Ming Chiao Tung University \\
  \texttt{yulunliu@cs.nycu.edu.tw} \\
}

\iclrfinalcopy 
\begin{document}

\maketitle
\begin{center}
  \vspace{-1.5em}
  \textbf{Project page:} \url{https://chien90190.github.io/worldhmr-fastcam-website/}
\end{center}
\begin{center}
    \includegraphics[width=\linewidth]{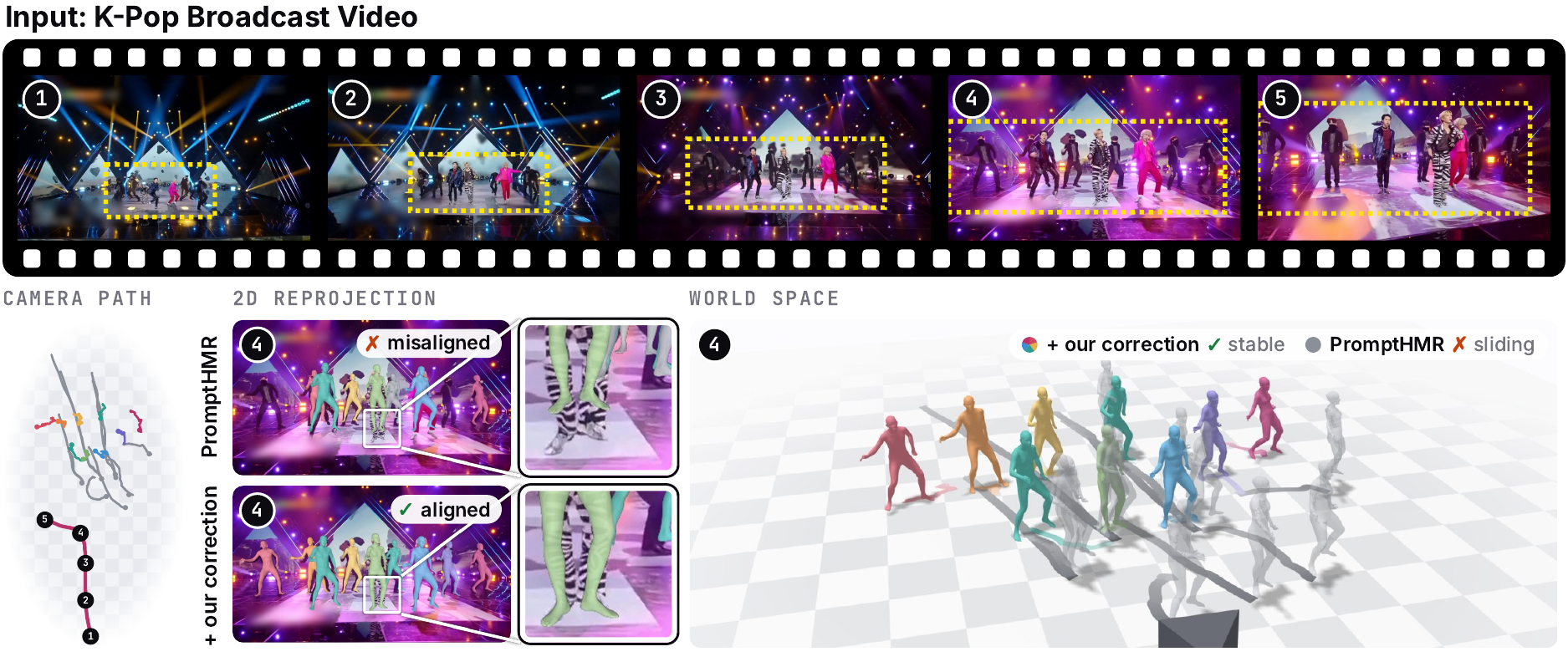}
    \captionof{figure}{\small \textbf{World-space drift under high camera dynamics takes the form of a scale mismatch between camera and human translation.} As the camera pushes toward the K-Pop performers \emph{(top, bottom left)}, PromptHMR~\citep{wang2025prompthmr} produces misaligned image-space reprojection \emph{(bottom middle)} and world-space trajectories that drift severely toward the camera \emph{(gray, bottom right)}. We trace this drift to how camera and human translations are combined. Each is recovered at its own scale, so a mismatch between the two lets camera motion contaminate human trajectories. Applied to existing backends, our closed-form correction estimates this mismatch and recovers stable trajectories and accurate reprojection \emph{(colored)}, with performers staying close to their stage positions.}
    \label{fig:teaser_new}
\end{center}

\begin{abstract}
    Monocular world-space human recovery methods excel on current benchmarks, but these rarely include fast camera motion. We introduce the K-Pop stress test, broadcast dance videos that combine fast, large camera motion with dense multi-person choreography. Leading methods produce plausible per-frame meshes on these videos, yet their world-space trajectories drift far from the performers' stage positions. Through a systematic diagnosis, we show that the camera and human estimates each remain self-consistent, and that the drift appears when they are combined at mismatched scales. Each is recovered at its own scale, so a mismatch between the two can turn camera motion into human drift that grows with camera displacement and becomes severe in broadcast dance videos. We derive a closed-form correction that estimates this mismatch and applies one scale to all performers, removing the drift while keeping their relative positions in one shared world frame. It requires no retraining and works with existing backends. Combined with physical constraints, the correction yields stable world-space reconstruction for choreography modeling, scene editing, data generation, and virtual reshooting.
\end{abstract}

\begin{figure}
    \centering
    \includegraphics[width=\linewidth]{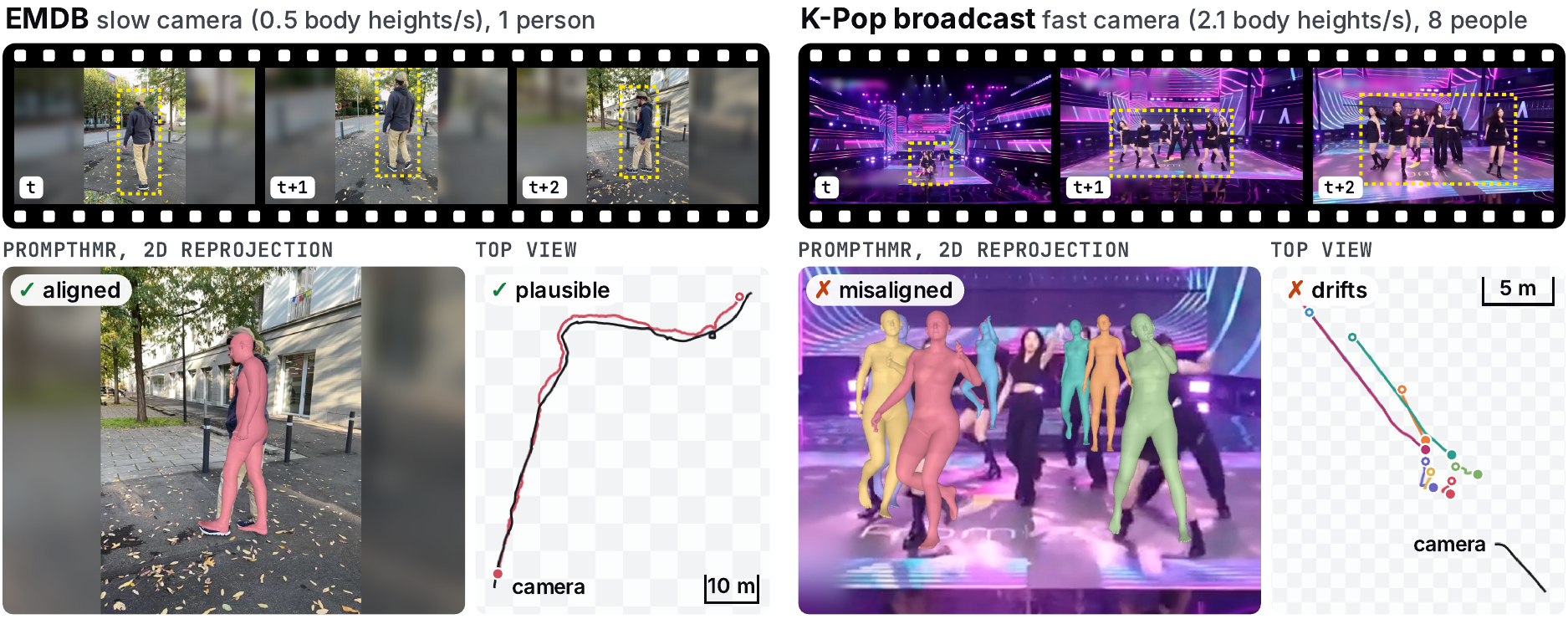}
    \caption{\textbf{World-space recovery that succeeds on existing benchmarks fails under high camera dynamics.} On EMDB~\citep{kaufmann2023emdb} \emph{(left)}, PromptHMR~\citep{wang2025prompthmr} recovers an accurate trajectory with aligned reprojection. On K-Pop broadcast video \emph{(right)}, its trajectories drift in the top view, and its meshes misalign in the reprojection. As the yellow boxes show, apparent size stays nearly constant on EMDB but changes rapidly under broadcast camera motion. Existing benchmarks seldom cover this regime, let alone combined with dense multi-person choreography.}
    \label{fig:motivation}
\end{figure}

\section{Introduction}
Recovering humans in world space~\citep{yuan2022glamr,shin2024wham,sun2023trace,shen2024world,wang2025prompthmr} from video is a central problem in computer vision, with applications in motion analysis, animation, and data generation. It requires estimating both human motion and camera motion. Recent approaches follow two main paradigms. The first directly predicts world-space motion from features such as pose and motion dynamics~\citep{yuan2022glamr,sun2023trace,shin2024wham,li2022d,shen2024world}. The second is compositional, combining estimated camera poses with camera-space human positions through a rigid transform~\citep{wang2024tram,wang2025prompthmr}. Supported by strong visual odometry and human mesh recovery models, compositional methods perform well on existing benchmarks, including 3DPW~\citep{von2018recovering}, RICH~\citep{huang2022capturing}, and EMDB~\citep{kaufmann2023emdb}, which span diverse human motions but mostly moderate camera motion (Figure~\ref{fig:camera_profile}).

Given this success, we ask whether compositional methods generalize to in-the-wild video beyond existing benchmarks. Commercial broadcast video, such as K-Pop performances, is one such setting, characterized by high camera dynamics, wide framing, and dense multi-person choreography. Recovering accurate world-space humans from such video would enable choreography modeling, scene editing, virtual reshooting, and data collection for training end-to-end methods. Surprisingly, when we test compositional methods on various K-Pop broadcast videos, performer trajectories drift severely in world space and become misaligned in image space (Figure~\ref{fig:motivation}).

To localize this failure, we first ask whether the visual odometry or the human mesh recovery model fails. We find that camera estimation remains self-consistent under a time-reversal test~\citep{huang2025vipe}, and that camera-space pose estimates remain visually plausible across frames. With both passing, we turn to the step that joins them. In monocular video, camera translation and camera-space human position are each recovered up to their own scale, yet composition treats the two scales as equal. We hypothesize that a mismatch between them makes camera motion appear as world-space human motion, causing the drift. Such a mismatch would also explain why both checks pass, since each estimate stays self-consistent under any scale shared across the clip.

\begin{figure}
    \centering
    \includegraphics[width=\linewidth]{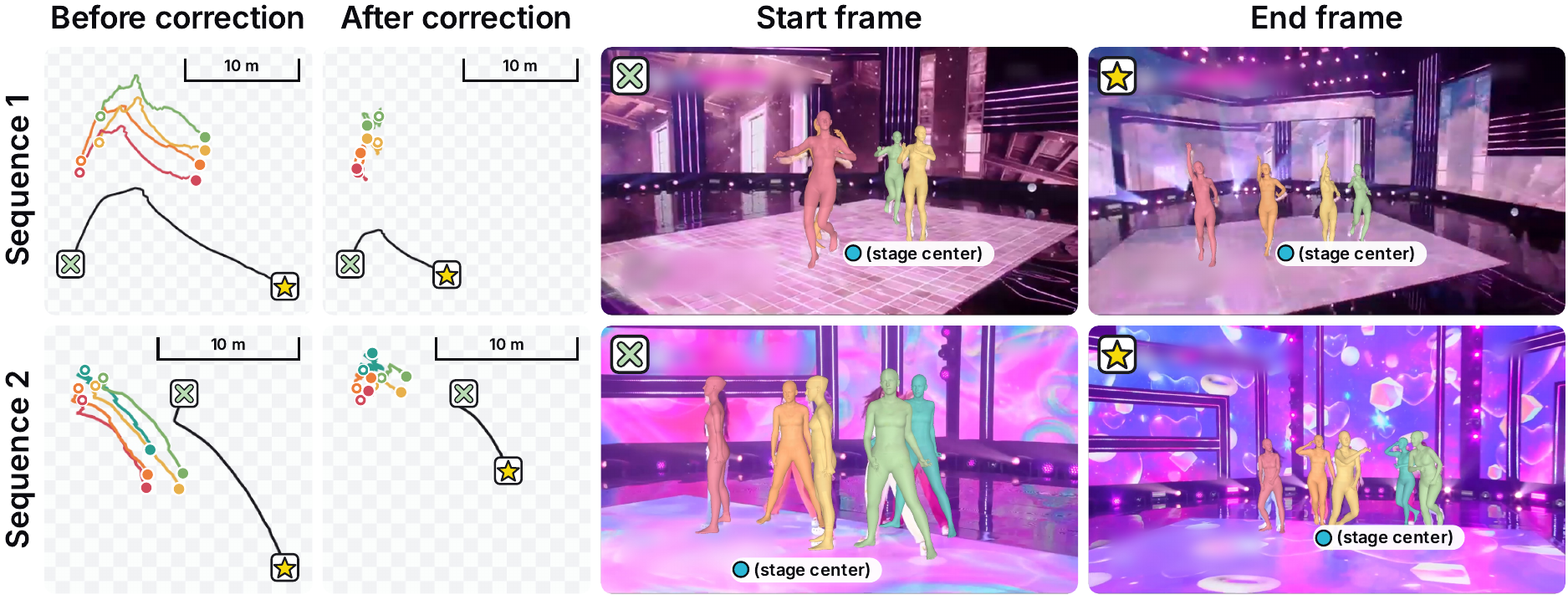}
    \caption{\textbf{Performer trajectories before and after drift correction.} \textit{Left:} World-space performer trajectories (colored) and the camera path (black) in the $x$-$z$ plane, with the camera's start (crossed box) and end (star). Before correction, the trajectories follow the camera's sweep; after, they do not. Tiles are 1~m. \textit{Right:} Start and end frames, with the stage center marked. Camera-space mesh overlays stay plausible despite the world-space drift.}
    \label{fig:drift_visual}
\end{figure}

To test this hypothesis without world-space ground truth, which broadcast video lacks, we examine intervals where performers remain visibly stationary. Their true world-space displacement is ideally zero, so any recovered displacement reflects the composition rather than performer motion. In these intervals across K-Pop broadcast videos, we observe drift that follows the camera path (Figure~\ref{fig:drift_visual}), supporting our hypothesis. We further find that this drift grows with camera displacement, which is why it stays small under the moderate camera motion of existing benchmarks.

Because the drift and the camera displacement are both observable in these stationary intervals, we solve for the scale mismatch in closed form. The resulting correction applies one scale to the camera and all performers, removing camera-induced drift while preserving the performers' relative positions in one shared world frame. It requires no retraining and transfers across backbones. Combined with physically grounded optimization, our method yields stable world-space reconstruction and accurate image-space reprojection for all performers (Figure~\ref{fig:teaser_new}). Against fixed venue cameras that recorded the same performances, our method recovers the closest world scale.

Our contributions are threefold. (1) We introduce the K-Pop stress test, a set of broadcast dance videos with high camera dynamics and dense choreography that existing benchmarks do not cover. (2) We show that compositional methods fail in this regime through a scale mismatch between independently estimated camera and human translations. (3) We derive a closed-form correction that removes this mismatch across backbones without retraining.

\section{Related Work}
\label{sec:related}

\paragraph{From camera-space to world-space human motion.} World-space recovery emerged as the central open problem following video-based HMR. One line infers world-space motion from local cues such as gait and pose dynamics~\citep{yuan2022glamr,sun2023trace,shin2024wham,li2022d}, but this \emph{local-to-global} mapping is inherently ambiguous when motion priors do not match the scene, for example a subject dancing in place. A second line jointly optimizes camera trajectory, human motion, and scale via SLAM~\citep{ye2023decoupling,kocabas2024pace,henning2022bodyslam,henning2023bodyslam++}. Recent work couples gravity-aware coordinates or absolute depth with camera estimation~\citep{shen2024world,yin2024whac,yao2023w}, and TRAM~\citep{wang2024tram} composes camera-space humans into a metric world frame using masked DROID-SLAM~\citep{teed2021droid} with metric scene depth, a pattern followed at larger scale by PromptHMR~\citep{wang2025prompthmr}. We follow this staged-composition family but identify the \emph{camera--human scale mismatch} as the dominant failure mode under high camera dynamics, a regime where these methods systematically degrade despite producing plausible camera-space meshes on existing benchmarks~\citep{von2018recovering,kaufmann2023emdb,huang2022capturing}. For more related work, see Appendix~\ref{app:related_work}.

\section{Background and Existing Gap}
\label{sec:background}

\textbf{Human mesh recovery (HMR).}
An HMR model estimates a parametric 3D human body from image observations, typically using SMPL~\citep{loper2023smpl} or its expressive extension SMPL-X~\citep{pavlakos2019expressive}, which map shape and pose parameters to a dense mesh. Given a video, the model predicts these parameters per frame along with a root translation that places the pelvis in camera space, a coordinate frame that moves with the camera. Recent methods compose these camera-space estimates with visual odometry to recover world-space motion~\citep{wang2024tram, wang2025prompthmr}. We adopt and study this compositional paradigm.

\textbf{Visual odometry (VO).}
Given a monocular video, visual odometry estimates camera motion from image observations, producing a sequence of camera poses $(R_i, T_i) \in SO(3) \times \mathbb{R}^3$, where $R_i$ and $T_i$ denote the camera-to-world rotation and translation at frame $i$ relative to a fixed world frame. Because the video is monocular, the recovered translation is defined only up to an unknown scale.

\textbf{World-space composition.}
Recovering humans in world space from monocular video combines the above two independent estimates through a rigid transform,
\begin{equation}
W_i = R_i \mathbf{X}_i + T_i,
\label{eq:composition}
\end{equation}
where $\mathbf{X}_i \in \mathbb{R}^3$ is the camera-space position of a tracked body point. We track the feet in our analysis because they remain in contact with the ground while the pelvis can move.

\textbf{Existing benchmarks lack high camera dynamics.}
Benchmarks for world-space human recovery, including 3DPW~\citep{von2018recovering}, RICH~\citep{huang2022capturing}, and EMDB~\citep{kaufmann2023emdb}, provide ground-truth trajectories from IMUs, electromagnetic sensors, or multi-view rigs, but primarily contain everyday, moderate camera motion. Synthetic benchmarks~\citep{patel2021agora,black2023bedlam,tesch2025bedlam2} expand body and appearance variation but rely on simulated camera trajectories that do not capture the large displacements and rapid motion changes of broadcast video. Therefore, existing benchmarks leave the robustness of Eq.~\ref{eq:composition} under sustained high camera dynamics largely untested. We address this gap with the K-Pop stress test in Section~\ref{sec:kpop_test}.

\begin{figure}
    \centering
    {(a) Camera profile statistics\par}
    {%
    \footnotesize
    \resizebox{\linewidth}{!}{%
    \tiny
    \begin{tabular}{l c c c l}
    \toprule
    \textbf{Metric} & \textbf{K-Pop} & \textbf{EMDB} & \textbf{WorldPose} & \textbf{Implication} \\
    \midrule
    Speed (body/s) & 1.15 & 0.20 & 0.14 & Higher average camera motion. \\
    Max speed (body/s) & 2.84 & 0.48 & 0.34 & More extreme motion bursts. \\
    Alignment (deg) & 3.64 & 9.48 & 6.02 & Tighter subject centering. \\
    Distance (body) & 4.88 & 1.58 & 27.48 & Wider framing than single-person capture. \\
    Acceleration/speed (1/s) & 1.94 & 1.10 & 0.96 & More abrupt motion changes relative to speed. \\
    \bottomrule
    \end{tabular}%
    }
    }
    \vspace{2mm}
    \newline
    {(b) Dataset statistics\par}
    \includegraphics[width=\textwidth]{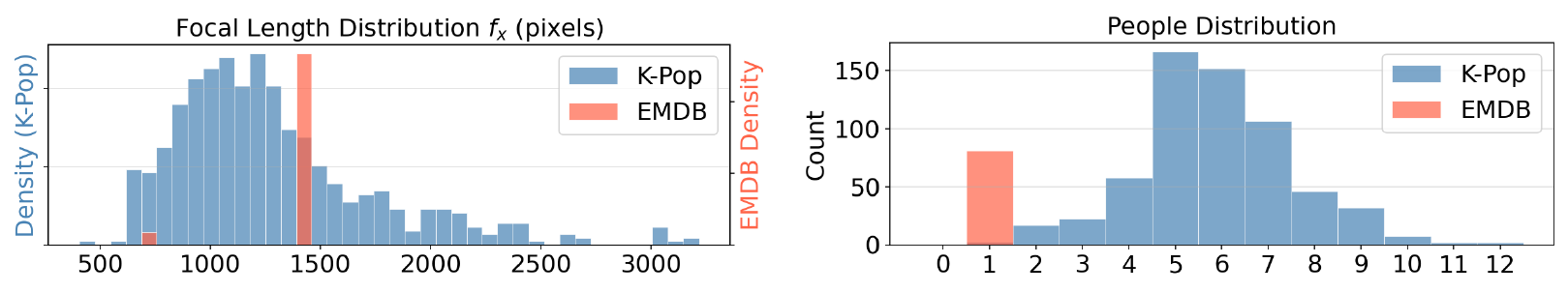}
    \caption{\textbf{The K-Pop stress test has faster, more abrupt camera motion than EMDB~\citep{kaufmann2023emdb} and WorldPose~\citep{jiang2024worldpose}, and denser scenes than EMDB.} (a) K-Pop cameras move about 6 to 8 times faster on average. The unit \emph{body} is one body length, making the three sets comparable despite different framing. (b) The stress test spans a broader focal-length range and up to 12 performers per frame, against EMDB's near-single-person clips.}
    \label{fig:camera_profile}
\end{figure}
\section{The K-Pop Stress Test}
\label{sec:kpop_test}
To evaluate world-space human recovery under high camera dynamics and dense multi-person choreography, we construct the K-Pop stress test from broadcast dance videos. We select publicly available clips with sufficient performer visibility and keep those whose camera trajectories pass the time-reversal test of Section~\ref{sec:problem_analysis}, so that every clip starts from a self-consistent camera estimate (Appendix~\ref{app:dataset_construction}). The resulting 616 clips contain 59.19k frames and 336.87k pose instances with up to 12 performers per frame (Figure~\ref{fig:camera_profile}, Appendix Table~\ref{sup-tab:dataset_comparison}), and we use them only for evaluation.

\textbf{Paired-camera reference set.}
Broadcast video provides no world-space ground truth. For direct evidence, we build a reference from performances also filmed by a fixed venue camera. The venue camera barely moves and keeps the whole group in view, so its reconstruction, which we obtain with PromptHMR~\citep{wang2025prompthmr}, is largely free of camera-induced drift and covers every performer. We hand-match 11 time-synchronized pairs covering 989 frames and 78 performers, over which the broadcast camera translates, zooms, and crops performers out of frame (Figure~\ref{fig:reference_set}). We report Ratio, the single scale that best fits a method's world-space positions to the reference over the clip (1 is perfect, and below 1 means too large), alongside WA-MPJPE, W-MPJPE, and PA-MPJPE. Pair Distance (the mean error in inter-performer distance) and Formation Error (the RMSE of the floor-plane arrangement) measure relative geometry alone, testing the shared world frame.

\textbf{Proxy metrics.}
On all 616 clips without reference, we measure reprojection against 2D keypoints from YOLO11n-Pose~\citep{yolo11_ultralytics} and masks from Mask2Former~\citep{cheng2021mask2former}, independent of the ViTPose~\citep{xu2022vitpose} detections our optimization fits. We report PCK (keypoints within 15\% of body height) and Mask IoU. For world-space stability, we report Jitter (temporal instability of joint positions from third-order finite differences) and Foot Skating (per-frame displacement of foot joints while feet are within 50\,mm of the ground plane).

\begin{figure}
    \centering
    \includegraphics[width=\linewidth]{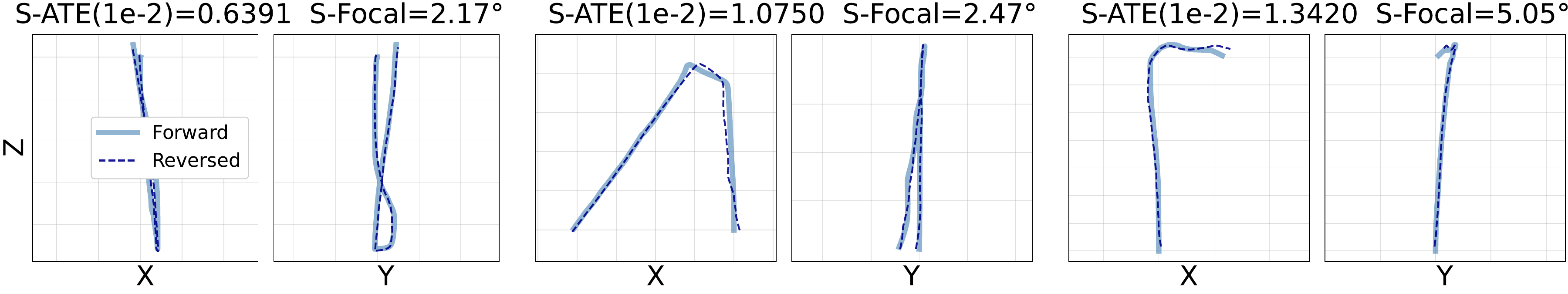}
    \caption{\textbf{ViPE~\citep{huang2025vipe} camera trajectories on K-Pop clips remain self-consistent under the time-reversal shuttle test.} Each clip is shown as a top-down view on the ground plane ($x$ against $z$) and a side view against vertical height ($y$ against $z$). The solid line is the forward pass and the dashed line the reversed pass, after a single similarity alignment. The rightmost panel is near at the K-Pop mean S-ATE of 1.35.}
    \label{fig:shuttle}
\end{figure}

\section{Diagnostic Analysis of Compositional Methods}
\label{sec:problem_analysis}
\begin{figure}[t]
    \centering
    \includegraphics[width=\linewidth]{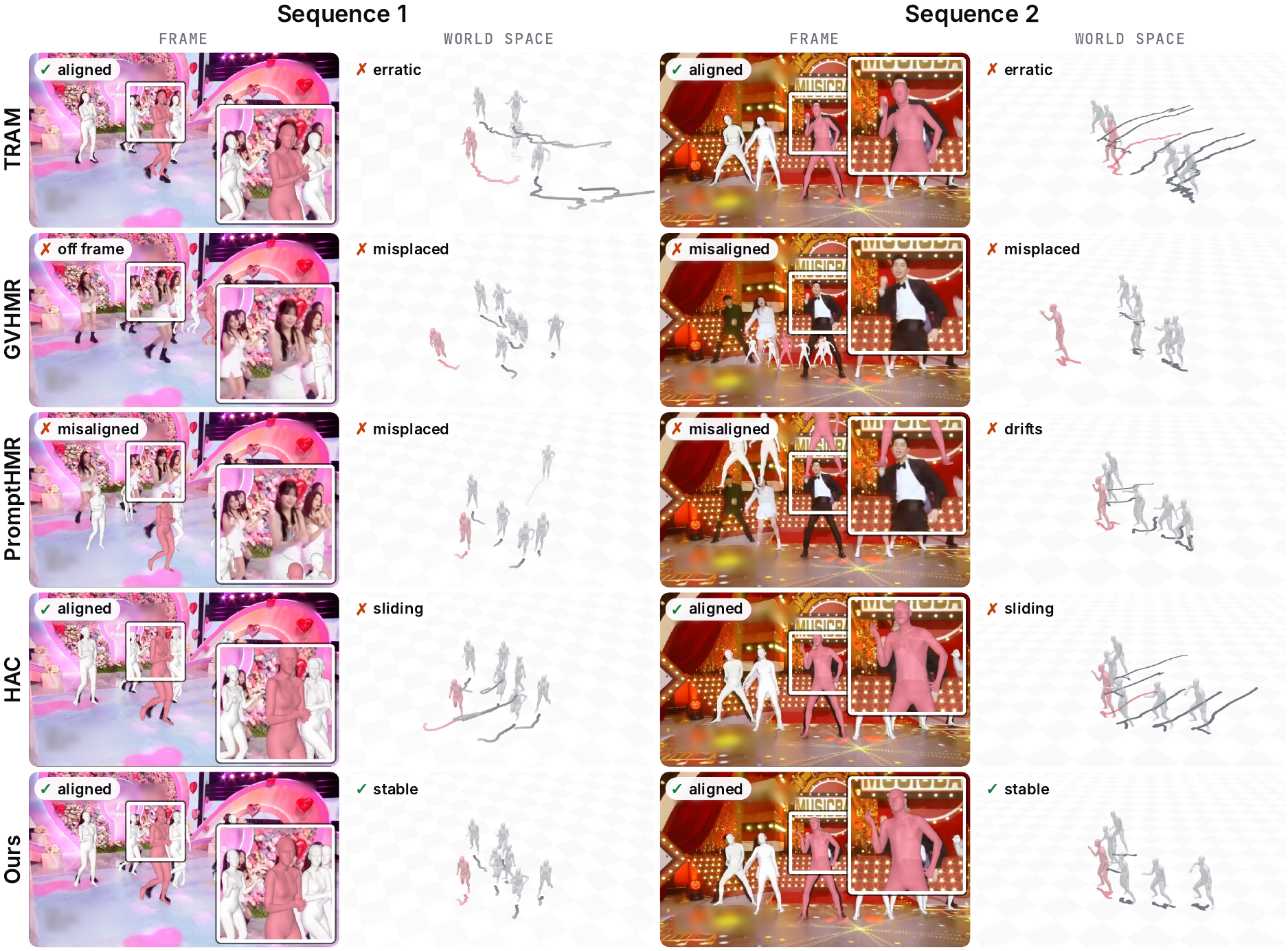}
    \caption{\textbf{Existing methods fail at 2D alignment, world-space stability, or both, while ours achieves both.} Each method shows mesh reprojection, with a close-up, and world space with each performer's path. One performer is in red. Tiles are 1~m. TRAM~\citep{wang2024tram} has erratic trajectories. GVHMR~\citep{shen2024world} misplaces performers in world space, and its reprojection leaves the frame or misaligns. PromptHMR~\citep{wang2025prompthmr} misaligns, misplaces, and drifts. HAC~\citep{yang2025humans} aligns but slides. Ours keeps both. Video comparisons: \url{https://chien90190.github.io/worldhmr-fastcam-website/}.}
    \label{fig:main_comparison}
\end{figure}

We run state-of-the-art compositional methods~\citep{wang2024tram, shen2024world, wang2025prompthmr} on the stress test. They exhibit severe world-space drift that follows camera ego-motion rather than random noise or occasional estimation error (Figure~\ref{fig:main_comparison}, Figure~\ref{fig:drift_visual}, left; Table~\ref{tab:exp_comparison}), suggesting a structural cause. World space is recovered by composing two estimates (Eq.~\ref{eq:composition}), so we examine each estimate alone and then their composition.

\textbf{(1) Camera-space HMR examination.}
One explanation is that the HMR model becomes unreliable under broadcast conditions, where performers are small in frame, densely arranged, and rapidly changing in apparent size. On K-Pop broadcast video, however, camera-space meshes remain plausible across frames, even during rapid reframing and large apparent-size changes (Figure~\ref{fig:drift_visual}), consistent with the strong benchmark performance of recent video-based HMR methods (Table~\ref{tab:hmr_quality}). We observe no frame-to-frame failure of camera-space HMR in these overlays.

\textbf{(2) Visual odometry examination.}
A second explanation is that visual odometry degrades under fast motion, changing backgrounds, and frequent reframing of broadcast video. Without ground-truth trajectories, we measure pose quality with a time-reversal shuttle test~\citep{huang2025vipe}, processing each clip forward and backward and aligning the two trajectories with a single similarity transform. Shuttle Absolute Trajectory Error (S-ATE, $\times 10^{-2}$, normalized by trajectory length) measures the root-mean-square positional deviation between them, and Shuttle Focal Error (S-Focal) measures the field-of-view discrepancy in degrees between the two estimated focal lengths. We use OpenDV, a ViPE test set of smooth, forward-dominant dashcam video, as the reference for accepted pose quality, and keep only K-Pop clips whose shuttle errors are comparable to it (Appendix~\ref{app:dataset_construction}). The final clips reach an S-ATE of 1.35 and an S-Focal of 3.39 degrees, against 1.05 and 4.26 on OpenDV (Figure~\ref{fig:shuttle}), so their camera estimates are self-consistent.

\begin{figure}
    \centering
    \includegraphics[width=\linewidth]{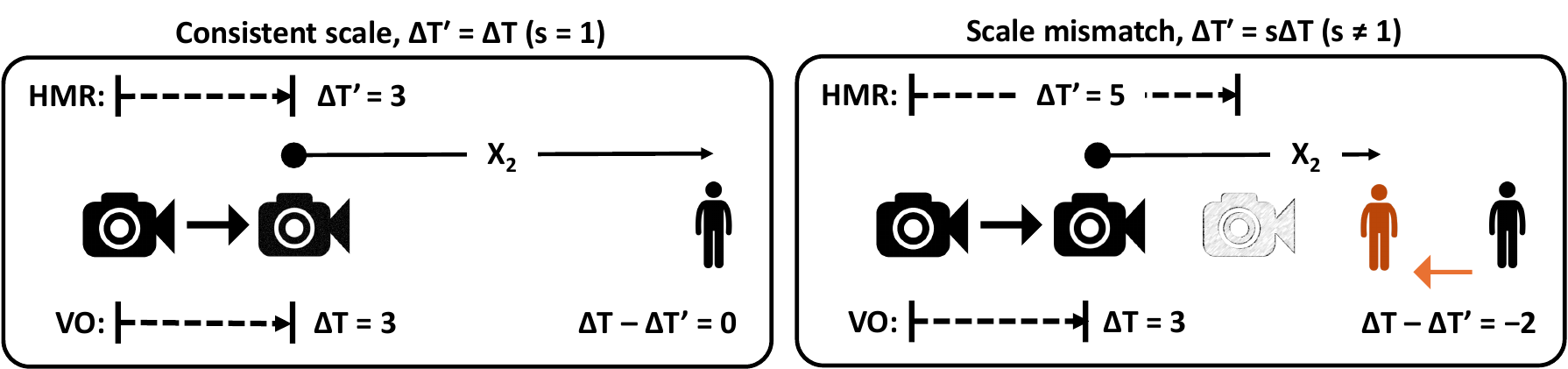}
    \caption{\textbf{Scale mismatch turns camera motion into world-space drift.} VO and HMR give two estimates of the same camera translation, $\Delta T$ and $\Delta T'$ (Eq.~\ref{eq:decomp}). If they agree, the static performer stays in place \emph{(left)}. If $\Delta T' = s\Delta T$ with $s \neq 1$, the performer drifts by $(1-s)\Delta T$ \emph{(right)}. HMR sees the performer's distance shrink by 5, but VO moves the camera by only 3. To keep the 5-unit shrink, composition moves the static performer to the orange position, 2 units closer to the camera. $\mathbf{X}_2$ is the performer's camera-space position after the camera moves.}
    \label{fig:cartoon_drift}
\end{figure}

\textbf{(3) Composition examination.}
Neither test can see an error shared across a whole clip. A depth bias common to all frames leaves mesh overlays unchanged, and a scale error shared by the forward and backward passes leaves them consistent. Such an error would surface only when the two estimates are composed, since the camera translation $T_i$ and camera-space position $\mathbf{X}_i$ are estimated independently and their scales can disagree. For a body point $p$ between frames $i$ and $j$, Eq.~\ref{eq:composition} gives
\begin{equation}
\Delta\hat{W}_{p,ij} = R_i\mathbf{X}_{p,i} - R_j\mathbf{X}_{p,j} + \Delta T_{ij},
\qquad
\Delta T_{ij}=T_i-T_j.
\label{eq:composed_displacement}
\end{equation}
The term $R_i\mathbf{X}_{p,i}$ is the performer's position relative to the camera in world space, so its change combines the performer's own world-space motion $\Delta W_{\mathrm{obj},p,ij}$ with the camera displacement implied by the camera-space estimates, $\Delta T'_{ij}$. Writing $R_i\mathbf{X}_{p,i} - R_j\mathbf{X}_{p,j} = \Delta W_{\mathrm{obj},p,ij} - \Delta T'_{ij}$ gives
\begin{equation}
\Delta\hat{W}_{p,ij} = \Delta W_{\mathrm{obj},p,ij} + \Delta T_{ij} - \Delta T'_{ij},
\label{eq:decomp}
\end{equation}
which holds for any camera rotation, with errors in the estimated rotation handled by the interval gates of Section~\ref{sec:method}. Here, visual odometry and HMR give two estimates of the same camera motion. When their scales agree, $\Delta T_{ij}$ and $\Delta T'_{ij}$ cancel, and a mismatch leaves camera-induced drift (Figure~\ref{fig:cartoon_drift}). Eq.~\ref{eq:decomp} alone cannot separate this mismatch from performer motion, and broadcast video has no world-space ground truth. We therefore turn to stationary intervals, where $\Delta W_{\mathrm{obj},p,ij}\approx 0$. Modeling the mismatch as $\Delta T'_{ij} = s\,\Delta T_{ij}$, where $s=1$ denotes consistent scale, we get
\begin{equation}
\Delta\hat{W}_{p,ij} \approx (1-s)\,\Delta T_{ij}.
\label{eq:stationary_drift}
\end{equation}
This drift grows with camera displacement, so it stays small on existing benchmarks and becomes severe under broadcast camera work. Focal-length error and monocular depth ambiguity can produce the mismatch by affecting the two estimates differently (Figure~\ref{fig:focal_ablation}, Appendix~\ref{app:other_sources}). Prior work estimates this scale from background scene depth~\citep{wang2024tram} or absolute HMR joint depth~\citep{yang2025humans,yang2024ofcam}. Both sides of Eq.~\ref{eq:stationary_drift} are measurable, so we estimate $s$ from relative displacements instead of absolute depth (Section~\ref{sec:method}).

\textbf{Reprojection misalignment from drift.} The scale mismatch explains the world-space drift, but not the observed image-space misalignment. Eq.~\ref{eq:composition} itself introduces no reprojection error because it can be exactly inverted using the same estimated camera pose. Misalignment instead arises during downstream optimization used by some pipelines to enforce physical plausibility. Large initial drift requires larger corrections, and the resulting non-convex optimization can converge to a physically plausible trajectory that is poorly aligned with the image. PromptHMR~\citep{wang2025prompthmr} performs such optimization and can therefore convert world-space drift into reprojection misalignment. TRAM~\citep{wang2024tram} does not and remains the unstable trajectory (Table~\ref{tab:exp_comparison}, Figure~\ref{fig:main_comparison}).

\section{Closed-Form Solution Derivation}
\label{sec:method}

Eq.~\ref{eq:stationary_drift} turns the diagnosis into an estimator, relating the observed world-space drift directly to the relative scale $s$. We estimate $s$ from stationary intervals by weighted least squares,
\begin{equation}
\min_s
\sum_{p,i,j}
w_{p,ij}
\left\|
(1-s)\Delta T_{ij}
-
\Delta\hat{W}_{p,ij}
\right\|^2,
\label{eq:scale_objective}
\end{equation}
where $w_{p,ij}$ reflects the confidence that the interval is stationary. The closed-form solution is
\begin{equation}
s^*
=
1-
\frac{
\sum_{p,i,j}
w_{p,ij}\,
\Delta\hat{W}_{p,ij}^{\top}\Delta T_{ij}
}{
\sum_{p,i,j}
w_{p,ij}\,
\|\Delta T_{ij}\|^2
}.
\label{eq:closed_form_solution}
\end{equation}

We apply the recovered scale to the visual odometry and world-space human translation,
\begin{equation}
W_i^{\mathrm{corr}}
=
R_i\mathbf{X}_i+s^*T_i
=
W_i-(1-s^*)T_i.
\label{eq:corrected_world}
\end{equation}
This removes camera-induced drift for all performers at once, preserving their relative positions in one shared world frame. We refine $s^*$ with a residual scale $\Delta s$ in the physically grounded optimization (PGO), which also enforces reprojection consistency, temporal smoothness, and inter-person separation (Appendix~\ref{app:losses}).

\begin{table}[t]
    \caption{\textbf{Ours is closest to the reference.} Methods run on the broadcast view and are compared against the fixed-camera reference. Tracked counts recovered performers, and Ratio is defined in Section~\ref{sec:kpop_test}. HAC~\citep{yang2025humans} shares our backbones. \colorbox{red!25}{Red} and \colorbox{orange!25}{orange} mark the best and second-best results.}
    \label{tab:paired_reference}
    \vspace{-2mm}
    \centering
    \footnotesize
    \resizebox{\linewidth}{!}{%
    \begin{tabular}{l c cc c cc c}
    \toprule
    \multirow{2}{*}{\textbf{Method}} & \textbf{Tracked} & \textbf{WA-MPJPE} & \textbf{W-MPJPE} & \textbf{PA-MPJPE} & \textbf{Pair dist.} & \textbf{Formation} & \textbf{Ratio} \\
    & (of 78) $\uparrow$ & (cm) $\downarrow$ & (cm) $\downarrow$ & (cm) $\downarrow$ & (cm) $\downarrow$ & (cm) $\downarrow$ & $\rightarrow 1$ \\
    \midrule
    TRAM~\citep{wang2024tram} & \cellcolor{red!25}\textbf{72} & 57.80 & 99.08 & 6.48 & 43.97 & \cellcolor{red!25}\textbf{30.28} & 0.77 \\
    PromptHMR~\citep{wang2025prompthmr} & \cellcolor{orange!25}66 & \cellcolor{orange!25}49.25 & 70.60 & 7.74 & \cellcolor{orange!25}42.07 & 34.67 & \cellcolor{orange!25}0.80 \\
    GVHMR~\citep{shen2024world} & 65 & 62.64 & \cellcolor{orange!25}68.16 & 8.76 & 50.67 & 53.72 & 0.64 \\
    HAC~\citep{yang2025humans} & \cellcolor{red!25}\textbf{72} & 64.77 & 116.30 & \cellcolor{orange!25}6.33 & 57.52 & 35.74 & 0.66 \\
    \midrule
    \textbf{Ours} & \cellcolor{red!25}\textbf{72} & \cellcolor{red!25}\textbf{39.89} & \cellcolor{red!25}\textbf{47.02} & \cellcolor{red!25}\textbf{6.20} & \cellcolor{red!25}\textbf{29.01} & \cellcolor{orange!25}32.97 & \cellcolor{red!25}\textbf{0.90} \\
    \bottomrule
    \end{tabular}
    }
\end{table}

\begin{figure}[t]
    \centering
    \includegraphics[width=\linewidth]{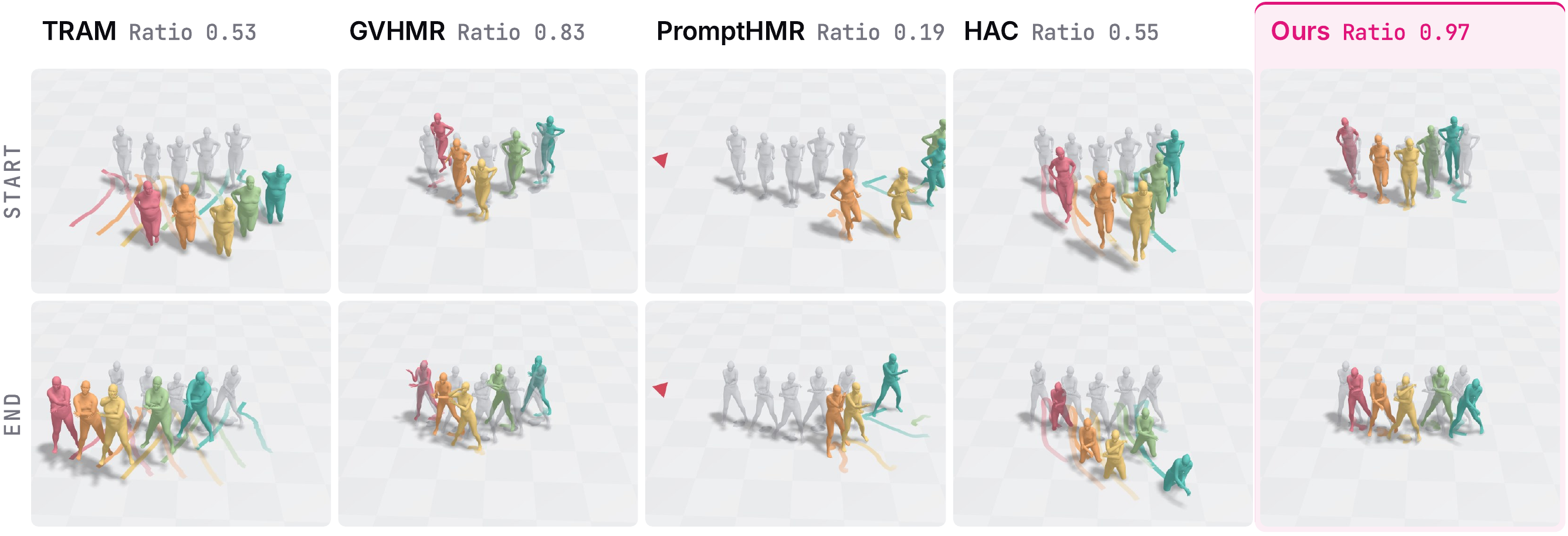}
    \caption{\textbf{A clip-level fit exposes sliding that single frames hide.} Sliding or a misplaced performer spreads a method's positions wider than the reference (gray), lowering its ratio even when one moment looks close. Rows show the first and last frames. TRAM~\citep{wang2024tram} and HAC~\citep{yang2025humans} look the right size but slide, and one PromptHMR~\citep{wang2025prompthmr} performer lands meters away (arrow). Tiles are 1~m.}
    \label{fig:ratio_visualize}
\end{figure}

\begin{table}
    \caption{\textbf{On the K-Pop stress test, ours is in the top two on both world-space metrics.} \colorbox{red!25}{Red}, \colorbox{orange!25}{orange}, and \colorbox{yellow!25}{yellow} mark the top three. TRAM~\citep{wang2024tram} and HAC~\citep{yang2025humans} only rigidly transform, so their reprojection cannot reveal wrong scales. HAC is furthest from the reference (Table~\ref{tab:paired_reference}). TRAM is unstable; HAC skates 2.8$\times$ more on the same backbones; GVHMR~\citep{shen2024world} reprojects poorly; PromptHMR~\citep{wang2025prompthmr} drifts severely.
    }
    \label{tab:exp_comparison}
    \centering
    \footnotesize
    \resizebox{0.83\linewidth}{!}{%
    \begin{tabular}{l cc cc}
    \toprule
    \multirow{2}{*}{\textbf{Method}} &
    \multicolumn{2}{c}{\textbf{2D Reprojection}} &
    \multicolumn{2}{c}{\textbf{3D World Space (mm)}} \\
    \cmidrule(lr){2-3}
    \cmidrule(lr){4-5}
    & \textbf{PCK \(\uparrow\)} & \textbf{Mask IoU \(\uparrow\)} &
    \textbf{Jitter \(\downarrow\)} & \textbf{Foot Skating \(\downarrow\)} \\
    \midrule
    TRAM~\citep{wang2024tram} & \cellcolor{red!25}0.921 & \cellcolor{red!25}0.848 & 312.39 & 131.29 \\
    GVHMR~\citep{shen2024world}     & 0.210 & 0.524 & \cellcolor{yellow!25}30.46 & \cellcolor{orange!25}31.57 \\
    PromptHMR~\citep{wang2025prompthmr} & 0.802 & 0.581 & $4.49\times10^{6}$ & $8.91\times10^{5}$ \\
    HAC~\citep{yang2025humans} & \cellcolor{orange!25}0.899 & \cellcolor{orange!25}0.747 & \cellcolor{red!25}17.56 & \cellcolor{yellow!25}50.48 \\
    \midrule
    \textbf{Ours} & \cellcolor{yellow!25}0.868 & \cellcolor{yellow!25}0.720 & \cellcolor{orange!25}26.87 & \cellcolor{red!25}17.94 \\
    \bottomrule
    \end{tabular}%
    }
\end{table}

\textbf{Stationary interval detection.}
Eq.~\ref{eq:closed_form_solution} requires negligible performer displacement. We identify candidate stationary intervals using foot-contact confidence and then reject intervals that show evidence of body motion or excessive rotational error. Recent HMR models~\citep{shen2024world,wang2025prompthmr} predict a per-joint contact confidence $c_{p,t,k}\in[0,1]$ for person $p$, joint $k$, and frame $t$. We define the strongest foot contact in each frame over the foot joints $\mathcal{C}$ as
\begin{equation}
\bar{c}_{p,t}
=
\max_{k\in\mathcal{C}}
c_{p,t,k}.
\end{equation}
We retain an interval $\mathcal{I}_{p,ij}$ when $\bar{c}_{p,t}$ exceeds $\tau_c=0.5$ in at least a fraction $\alpha=0.8$ of its frames. Foot contact alone does not guarantee that the performer is stationary, so we reject an interval if either (i) the person's displacement is not aligned with camera motion ($|\cos(\Delta\hat{W}_{p,ij},\Delta T_{ij})| \le 0.8$), since a stationary performer drifts only along the camera's path (Eq.~\ref{eq:stationary_drift}), or (ii) the expected rotational error dominates the translational signal, with a signal-to-noise ratio below 1.5 (Appendix~\ref{app:rotation}). Surviving intervals are weighted by their mean contact confidence,
\begin{equation}
w_{p,ij}
=
\frac{1}{|\mathcal{I}_{p,ij}|}
\sum_{t\in\mathcal{I}_{p,ij}}
\bar{c}_{p,t}.
\end{equation}
Because contact is predicted in camera space before composition, selecting stationary intervals does not depend on the drift from which Eq.~\ref{eq:closed_form_solution} recovers $s^*$. A backbone without contact prediction can instead obtain contact confidence from a separate model, as for TRAM in Table~\ref{tab:exp_backends} (Appendix~\ref{app:contact_input}).

\textbf{Multi-person collision.} Performers stand close together, so recovered meshes often interpenetrate. We penalize interpenetration and its rate of change with a temporal repulsion loss, which spreads corrections over nearby frames rather than a single frame (Appendix~\ref{app:losses}).

\textbf{Implementation.} Unless otherwise specified, we use PromptHMR-Vid~\citep{wang2025prompthmr} as the HMR backbone and ViPE~\citep{huang2025vipe} as the visual odometry. See Appendix~\ref{app:pipeline} for hyperparameter details. All experiments are conducted on a single NVIDIA RTX A6000 GPU.
\section{Experiments}
\label{sec:exp}
We compare against TRAM~\citep{wang2024tram}, GVHMR~\citep{shen2024world}, PromptHMR~\citep{wang2025prompthmr}, and HAC~\citep{yang2025humans}. PromptHMR and GVHMR predict metric-scale translation (Appendix~\ref{app:pipeline}). GVHMR is single-person and does not compose with visual odometry, so we extend it with multi-person detection and rigid alignment at each track's first frame. We run HAC, a scale estimator, on our backbones and average its per-person scales into one (Appendix~\ref{app:pipeline_step5}). The reference set tests scale directly (Table~\ref{tab:paired_reference}, Figure~\ref{fig:ratio_visualize}). Ours recovers a scale ratio closest to 1 and leads every baseline on WA-MPJPE, W-MPJPE, and Pair Distance, outperforming PromptHMR, which built the reference. On the full stress test, ours is the only method in the top two on both world-space metrics (Table~\ref{tab:exp_comparison}, Figure~\ref{fig:main_comparison}). Reprojection cannot reveal a wrong scale (Section~\ref{sec:problem_analysis}). HAC, like TRAM, skips optimization and reprojects well while placing performers furthest from the reference. Table~\ref{tab:exp_backends} shows gains across visual-odometry and HMR backbones.

\subsection{Ablation study}
\begin{figure}[t!]
    \centering
    {(a) PGO with vs. without drift correction on the 100 most dynamic K-Pop clips\par}
    {%
    \footnotesize
    \resizebox{0.65\linewidth}{!}{%
    \tiny
    \begin{tabular}{c cc cc}
    \toprule
    \multirow{2}{*}{\textbf{Drift Correction}} &
    \multicolumn{2}{c}{\textbf{2D Reprojection}} &
    \multicolumn{2}{c}{\textbf{3D World Space (mm)}} \\
    \cmidrule(lr){2-3}
    \cmidrule(lr){4-5}
    & \textbf{PCK \(\uparrow\)} & \textbf{Mask IoU \(\uparrow\)} &
    \textbf{Jitter \(\downarrow\)} & \textbf{Foot Skating \(\downarrow\)} \\
    \midrule
    No  & 0.727 & 0.641 & 32.17 & 31.08 \\
    Yes & \textbf{0.839} & \textbf{0.691} & \textbf{15.88} & \textbf{18.57} \\
    \bottomrule
    \end{tabular}%
    }
    }
    \includegraphics[width=\linewidth]{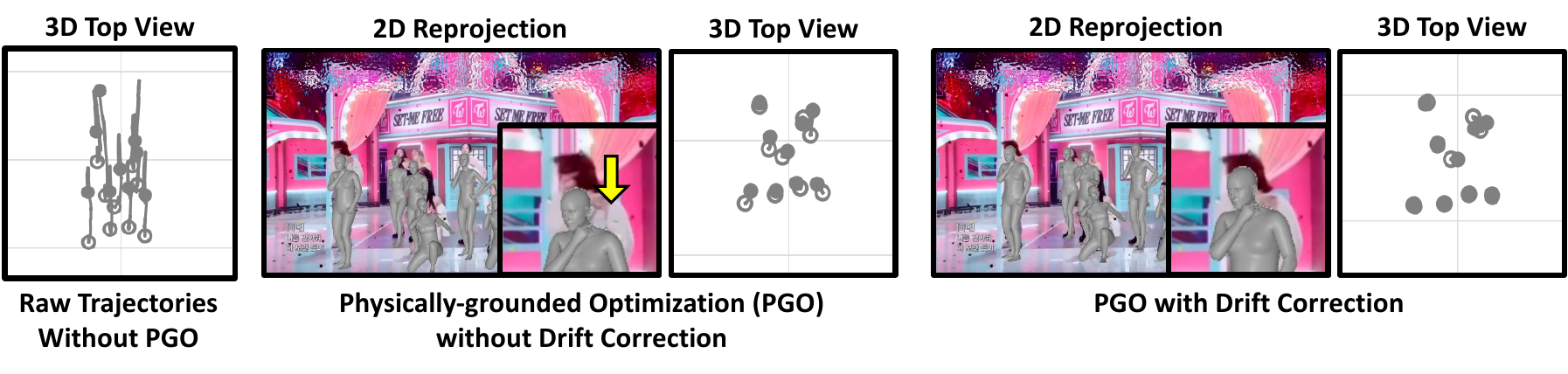}
    \newline
    {(b) Stationary performers under raw trajectories, PGO alone, and PGO with drift correction\par}
    \caption{\textbf{Drift correction is needed even with PGO.} (a) Adding drift correction improves every metric. (b) For stationary performers, each top-view trajectory should collapse to a point. Raw trajectories drift far \emph{(left)}. PGO partially shortens the drift but leaves meshes misaligned \emph{(middle)}. With drift correction, positions stay fixed, and meshes align \emph{(right)} (Appendix~\ref{app:stationary}).}
    \label{fig:ablation_drift}
\end{figure}

Drift correction improves both reprojection and world space across all metrics (Figure~\ref{fig:ablation_drift}(a), 100 most dynamic clips). PGO suppresses much of the drift on its own but leaves the scale mismatch in place, so stationary performers still creep across the scene until drift correction removes it (Figure~\ref{fig:ablation_drift}(b)). Temporal repulsion reduces inter-performer penetration and keeps bodies separated in multi-person sequences (Figure~\ref{fig:ablation_temporal}, Appendix). Across 8 sequences and 3 baselines, 34 participants preferred our method on overall quality and camera motion in over 96\% of comparisons (Appendix~\ref{app:user_study}).

\textbf{Moderate camera motion.} We evaluate on EMDB-2 to check that the correction does not severely degrade results under moderate camera motion. Against our reproduction of PromptHMR~\citep{wang2025prompthmr}, we obtain W-MPJPE 415.03 versus 388.66\,mm, WA-MPJPE 158.50 versus 139.65\,mm, Jitter 21.18 versus 46.12\,mm, and Foot Skating 8.79 versus 11.19\,mm. We attribute the weaker trajectory metrics to small camera displacement, which leaves Eq.~\ref{eq:closed_form_solution} little signal to estimate $s$ from, while both physical plausibility metrics improve.

\begin{table}[t!]
    \caption{\textbf{Drift correction transfers across visual odometry and HMR backbones.} Bold marks the better of each pair. $^\dagger$GVHMR in Table~\ref{tab:exp_comparison} keeps its own trajectory, aligned to DPVO~\citep{teed2023deep} once per track, so per-frame camera noise does not enter and its jitter there is lower. $^\ddagger$Contact confidence comes from PromptHMR-Vid, as TRAM predicts none. $^\S$TRAM~\citep{wang2024tram} only rigidly transforms camera-frame predictions, so reprojection returns the per-frame prediction, while its world-space motion inherits camera jitter and drift. On TRAM, our correction trades a small loss in 2D fit for a large gain in world-space stability.}
    \label{tab:exp_backends}
    \centering
    \footnotesize
    \resizebox{\linewidth}{!}{%
    \begin{tabular}{ll cc cc}
    \toprule
    \multirow{2}{*}{\textbf{HMR}} & \multirow{2}{*}{\textbf{Odometry}} & \multicolumn{2}{c}{\textbf{2D Reprojection}} & \multicolumn{2}{c}{\textbf{3D World Space (mm)}} \\
    \cmidrule(lr){3-4} \cmidrule(lr){5-6}
     & & \textbf{PCK ($\uparrow$)} & \textbf{Mask IoU ($\uparrow$)} & \textbf{Jitter ($\downarrow$)} & \textbf{Foot Skating ($\downarrow$)} \\
    \midrule
    \multicolumn{6}{l}{\textbf{Original methods with their native odometry}} \\
    PromptHMR~\citep{wang2025prompthmr} & \multirow{4}{*}{\shortstack[l]{masked DROID-SLAM\\\citep{teed2021droid}}} & 0.802 & 0.581 & $4.49\times10^{6}$ & $8.91\times10^{5}$ \\
    \quad + drift correction & & \textbf{0.859} & \textbf{0.720} & \textbf{27.53} & \textbf{18.65} \\
    \cmidrule(lr){1-1} \cmidrule(lr){3-6}
    TRAM~\citep{wang2024tram}$^\S$ & & \textbf{0.921} & \textbf{0.848} & 312.39 & 131.29 \\
    \quad + drift correction$^\ddagger$ & & 0.849 & 0.730 & \textbf{47.38} & \textbf{22.03} \\
    \midrule
    \multicolumn{6}{l}{\textbf{Other backbones with drift correction}} \\
    PromptHMR-Vid & \multirow{3}{*}{ViPE~\citep{huang2025vipe}} & 0.868 & 0.720 & 26.87 & 17.94 \\
    GVHMR & & 0.883 & 0.751 & 34.64$^\dagger$ & 22.89 \\
    TRAM$^\ddagger$ & & 0.858 & 0.738 & 46.92 & 20.75 \\
    \cmidrule(lr){1-6}
    PromptHMR-Vid & DA3~\citep{lin2025depth} & 0.868 & 0.706 & 26.88 & 18.22 \\
    \bottomrule
    \end{tabular}%
    }
\end{table}

\begin{figure}
    \centering
    \includegraphics[width=\linewidth]{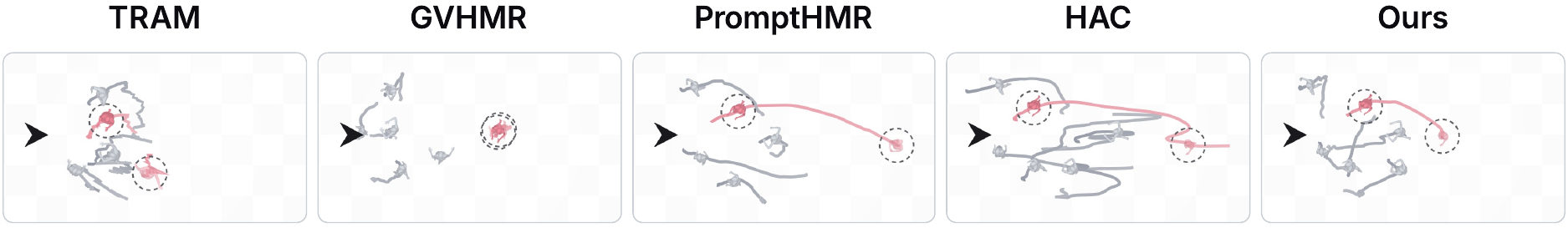}
    \caption{\textbf{Under long occlusion, our method places a performer meters behind the group, and the placement recovers once the occlusion ends.} Each panel is a top view with the camera at the arrow. The circled performer (red) is shown during and after the occlusion. All panels share one scale. Tiles are 1 m. PromptHMR~\citep{wang2025prompthmr} and HAC~\citep{yang2025humans} fail the same way. GVHMR~\citep{shen2024world} keeps the performer far. TRAM~\citep{wang2024tram} still jitters.}
    \label{fig:failure}
\end{figure}

\textbf{Failure case and limitations.} A single scale per sequence moves the group rigidly, so it cannot repair a performer whose camera-space depth is already wrong, such as the occluded performer in Figure~\ref{fig:failure}. Clips are already split at shot cuts (Appendix~\ref{app:dataset_construction}), but a scale that varies within a shot is not represented (Table~\ref{tab:scalemodel}). Scale recovery inherits noise from upstream camera and HMR estimates.

\section{Conclusion}
World-space human recovery appears robust on existing benchmarks, whose cameras mostly move mildly. Under the broadcast camera work of our K-Pop stress test, we show that a scale mismatch between camera and human translation turns camera motion into world-space drift. Our closed-form correction removes this drift without retraining, giving stable trajectories across backends, and we hope this motivates broader evaluation and application under high camera dynamics.


\bibliography{arxiv}
\bibliographystyle{arxiv}

\appendix
\section*{Appendix}
\begin{figure}[b]
    \centering
    \includegraphics[width=\textwidth]{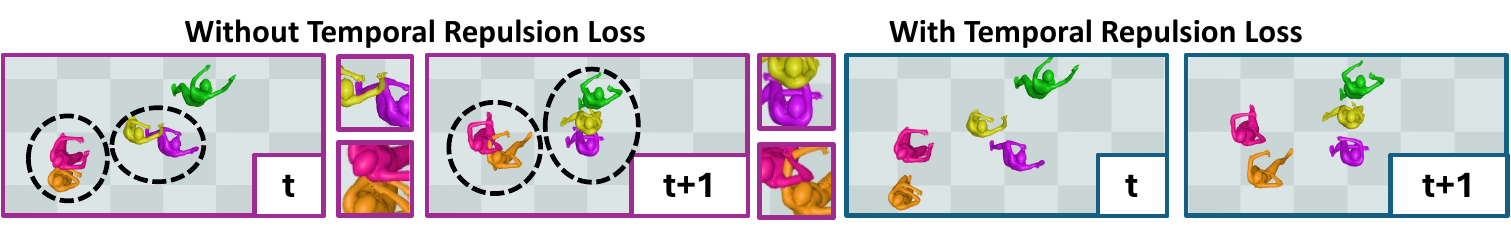}
    \caption{\textbf{Temporal repulsion improves multi-person consistency by preventing inter-performer collapse.} Without it, nearby performers penetrate and form physically implausible configurations that persist over time. The term maintains separation between bodies and produces a more coherent world-space motion across frames.}
    \label{fig:ablation_temporal}
    \vspace{-5mm}
\end{figure}

\begin{figure}
    \centering
    \includegraphics[width=\linewidth]{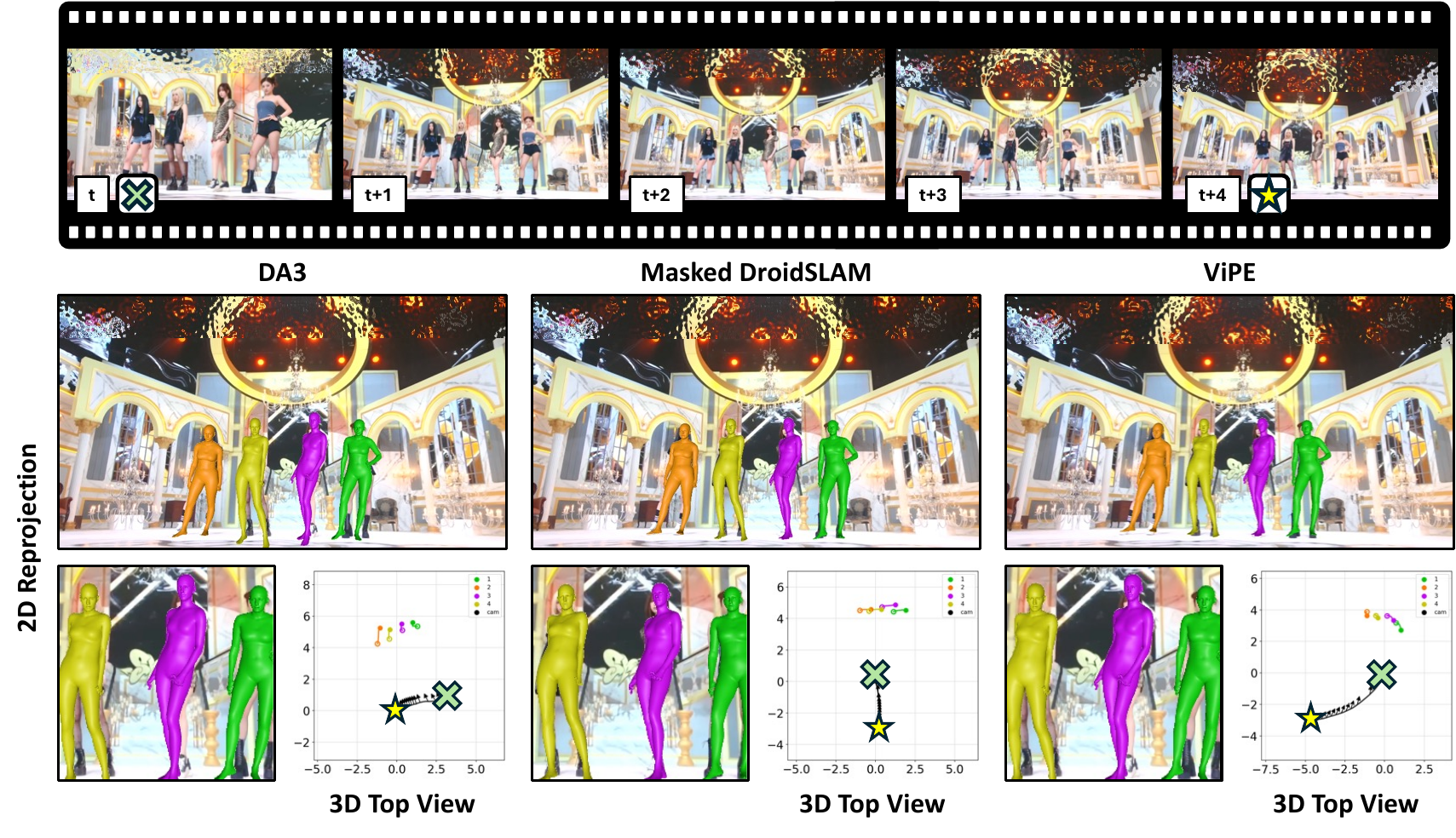}
    \caption{\textbf{Ablation of visual odometry.} All three methods produce plausible 2D reprojections with reasonable subject alignment, but differ in their estimated camera trajectories. Based on the input video without ground truth, the camera starts from the right, pulls back, and pans toward the center of the group. DA3 captures lateral movement but misses the pull-back, while masked DROID-SLAM shows minimal lateral movement. ViPE most accurately recovers this trajectory, estimating both the backward pull and the pan toward the group center.}
    \label{fig:sup-odometry}
\end{figure}

\begin{figure}
    \centering
    \includegraphics[width=\textwidth]{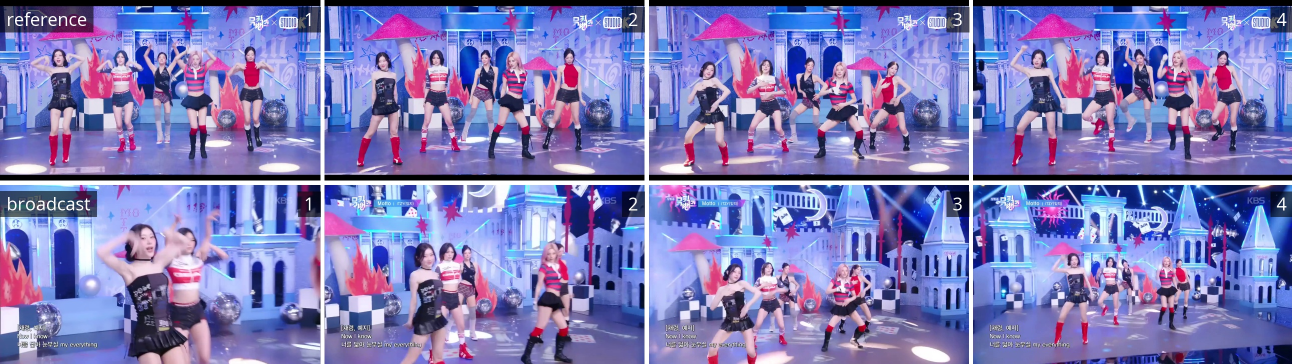}
    \caption{\textbf{Paired-camera reference set.} Four time-synchronized frames from one pair, numbered in the upper right. The fixed venue camera (top) holds its framing on the whole group, while the broadcast camera (bottom) translates, zooms, and crops performers out of frame over the same span.}
    \label{fig:reference_set}
\end{figure}

\textbf{Appendix overview.}
Appendix~\ref{app:related_work} expands the related work on human mesh recovery, world-space benchmarks, and physical priors. Appendices~\ref{app:pipeline} and~\ref{app:losses} give the implementation. The first shows each pipeline step from camera pose estimation to PGO, with an odometry ablation in Figure~\ref{fig:sup-odometry}. The second sets out the full loss formulation. Appendix~\ref{app:drift_assumption} states the assumptions behind the drift decomposition and the gates that enforce them. Appendix~\ref{app:scale_ablations} tests the estimator's two choices, sweeping the contact threshold (Table~\ref{tab:threshold}), replacing the single scale with a time-varying one (Table~\ref{tab:scalemodel}) and starving it of intervals (Table~\ref{tab:sparsity}), then checking the correction on clips where performers hold their positions (Table~\ref{tab:stationary}). Appendix~\ref{app:dataset_construction} describes how we built the K-Pop stress test and compares it against related benchmarks (Table~\ref{sup-tab:dataset_comparison}). Appendix~\ref{app:user_study} reports the perceptual user study (Figure~\ref{fig:user_study}). Figure~\ref{fig:ablation_temporal} shows the effect of the temporal repulsion term. Figure~\ref{fig:reference_set} illustrates the paired-camera reference set. Table~\ref{tab:hmr_quality} shows HMR performance on existing benchmarks.

\section{Related Work}
\label{app:related_work}
\paragraph{Human mesh recovery.}
Recovering parametric human bodies~\citep{loper2023smpl,pavlakos2019expressive} has progressed from optimization-based fitting~\citep{bogo2016keep} to end-to-end regression~\citep{kanazawa2018end,kolotouros2019learning}, with subsequent work targeting occlusion robustness~\citep{kocabas2021pare,choi2022learning,khirodkar2022occluded}, transformer-based mesh decoding~\citep{lin2021end,lin2021mesh,goel2023humans}, hybrid analytical inverse kinematics~\citep{li2021hybrik}, full-frame perspective reasoning~\citep{li2022cliff,SPEC:ICCV:2021,patel2025camerahmr,dwivedi2024tokenhmr}, and foundation-scale promptable inference~\citep{yang2026sam}. Multi-person extensions detect and regress all subjects in a single pass~\citep{sun2021monocular,sun2022putting,baradel2024multi,qiu2023psvt,wang2024multi}. Video-based extensions improved temporal smoothness in camera coordinates~\citep{kocabas2020vibe,choi2021beyond,wei2022capturing,shen2023global}, after which world-space recovery emerged as the central open problem, covered in the main paper.

\paragraph{Scale ambiguity in compositional recovery.}
The relative scale between independently estimated camera and human translation has become the explicit subject of recent work~\citep{wang2024tram,yang2025humans,yang2024ofcam,song2025metrichmsr}. HAC~\citep{yang2025humans} treats the body as a metric calibration object using absolute joint depth from an HMR network, and OfCaM~\citep{yang2024ofcam} similarly leverages HMR depth to bypass iterative optimization. Both produce a \emph{per-person} scale, which does not by itself give a shared world frame, so we average HAC's scales for comparison (Appendix~\ref{app:pipeline_step5}). MetricHMR~\citep{song2025metrichmsr} regresses metric translation from camera-ray maps but does not couple human and camera trajectories under dynamic ego-motion.

\paragraph{Benchmarks for world-space human recovery.}
Existing world-space benchmarks feature moderate, everyday camera motion. 3DPW~\citep{von2018recovering} captures outdoor handheld scenes with IMU ground truth, EMDB~\citep{kaufmann2023emdb} uses electromagnetic sensors to capture dynamic-camera trajectories, and RICH~\citep{huang2022capturing} adds scene contact in indoor multi-view settings. Synthetic benchmarks~\citep{patel2021agora,black2023bedlam,tesch2025bedlam2} expand body and appearance variation but under-represent the rapid translation, abrupt direction changes, and dolly-zooms of professional broadcast production. Dance and group-motion datasets cover controlled multi-view choreography~\citep{li2021ai}, in-the-wild groups without world-space pose~\citep{le2023music,sun2022dancetrack}, or animator-authored single-dancer sequences~\citep{wang2024dancecamera3d}. WorldPose~\citep{jiang2024worldpose} uses LiDAR and multi-view rigs for multi-person world-space motion and is closest to ours, but covers a sport-specific motion distribution. Human3.6M~\citep{ionescu2013human3}, MPI-INF-3DHP~\citep{mehta2017monocular}, and HumanML3D~\citep{guo2022generating} address camera- or motion-space evaluation and do not target world-space recovery. Our K-Pop stress test fills this gap by providing in-the-wild broadcast video with sustained \emph{high-magnitude} camera dynamics and dense multi-person choreography, exposing failure modes that current evaluation regimes leave understudied.

\paragraph{Camera estimation under dynamic conditions.}
Monocular SLAM~\citep{davison2007monoslam,klein2007parallel,engel2014lsd,engel2017direct,mur2015orb,mur2017orb,campos2021orb,forster2014svo} assumes a static scene and degrades under rapid lighting changes, motion blur, and non-rigid motion, while learning-based systems~\citep{teed2021droid,teed2023deep,lipson2024deep,wang2021tartanvo,wang2017deepvo} struggle when foreground subjects dominate. Recent work addresses this via learned masking~\citep{bescos2018dynaslam,zhao2022particlesfm,chen2024leap,rockwell2025dynamic}, dynamic-aware 3D models~\citep{zhang2024monst3r,li2025megasam}, or uncertainty-aware DROID-SLAM~\citep{li2026droid}.

\paragraph{Physical priors and foot--ground contact.}
Physical realism has been pursued via humanoid simulation and contact reasoning~\citep{shimada2020physcap,yuan2021simpoe,shimada2021neural,xie2021physics,feng2025physhmr}, generative motion priors~\citep{rempe2021humor,mahmood2019amass}, scene-contact constraints~\citep{hassan2019resolving,tripathi20233d,muller2021self}, and explicit foot-skating mitigation~\citep{zou2020reducing,rempe2020contact}. Foot--ground contact has been used as a kinematic constraint in trajectory optimization~\citep{rempe2020contact,shin2024wham,shen2024world} and, in concurrent work~\citep{yang2025humans}, as a calibration cue based on the \emph{absolute spatial position} of contact joints. Our use of contact differs in two respects. First, we leverage contact \emph{temporally}, since a stationary planted-foot interval is one over which the true human world displacement is approximately zero, isolating the camera-induced drift component. Second, the resulting constraint depends only on \emph{relative odometry translation}, not on absolute monocular depth, making it robust to the dolly-zoom dynamics and performative lighting that dominate broadcast video.

\begin{table}
    \caption{\textbf{Camera-space HMR methods perform strongly on existing benchmarks, consistent with the frame-to-frame plausibility we observe on K-Pop broadcast video.}
    Quantitative comparison of representative methods on 3DPW~\cite{von2018recovering}, RICH~\citep{huang2022capturing}, and EMDB-2~\citep{kaufmann2023emdb}, with the number of joints shown in parentheses. EMDB-2 results are in world space and are reported after alignment. All metrics are in millimeters. $^\ast$The 3DPW and RICH entries for PromptHMR-Vid~\citep{wang2025prompthmr} are those reported for the single-image PromptHMR, which is the variant its authors evaluate in camera space.}
    \label{tab:hmr_quality}
    \centering
    \footnotesize
    \setlength{\tabcolsep}{4pt}
    \resizebox{\linewidth}{!}{%
    \begin{tabular}{lccc ccc cccc}
    \toprule
    \multirow{2}{*}{\textbf{Method}} & \multicolumn{3}{c}{\textbf{3DPW (14)}} & \multicolumn{3}{c}{\textbf{RICH (24)}} & \multicolumn{4}{c}{\textbf{EMDB-2 (24)}} \\
    \cmidrule(lr){2-4}
    \cmidrule(lr){5-7}
    \cmidrule(lr){8-11}
    & PA-MPJPE & MPJPE & PVE & PA-MPJPE & MPJPE & PVE & WA-MPJPE100 & W-MPJPE100 & Jitter & Foot Skating \\
    \midrule
    WHAM~\citep{shin2024wham}          & 37.5 & 59.8 & 71.5 & 44.3 & 80.0 & 91.2 & 135.6 & 354.8 & 22.5 & 4.4 \\
    TRAM~\citep{wang2024tram}          & 35.6 & 59.3 & 69.6 & ---  & ---  & ---  & 76.4  & 222.4 & 18.5 & 23.4 \\
    GVHMR~\citep{shen2024world}         & 37.0 & 56.6 & 68.7 & 39.5 & 66.0 & 74.4 & 111.0 & 276.5 & 16.7 & 3.5 \\
    PromptHMR-Vid~\citep{wang2025prompthmr} & 35.5$^\ast$ & 56.9$^\ast$ & 67.3$^\ast$ & 37.0$^\ast$ & 57.4$^\ast$ & 65.8$^\ast$ & 71.0  & 216.5 & 16.3 & 3.5 \\
    \bottomrule
    \end{tabular}%
    }
\end{table}

\section{Pipeline Details}
\label{app:pipeline}
Our pipeline follows TRAM~\citep{wang2024tram} in separating camera estimation, camera-space human reconstruction, and world-space composition, but is adapted to the high-camera-dynamics regime studied in this work. Beyond recovering coherent world-space motion, a central goal is to expose and resolve the scale mismatch that becomes dominant under aggressive camera motion.

\paragraph{Step 1. Camera pose estimation.}
Standard dense SLAM methods such as DROID-SLAM~\citep{teed2021droid} are brittle on broadcast video with dynamic backgrounds, where large non-rigid image motion can corrupt feature matching and bundle adjustment, as shown in Figure~\ref{fig:sup-odometry}. We therefore use ViPE~\citep{huang2025vipe} with a constrained phrase list covering dynamic foreground elements (persons, dancers, singers, handheld props), broadcast overlays (lyrics, subtitles, captions), and background regions (sky) to reduce the influence of background motion on camera estimation, sampled at a keyframe gap of 0.1\,s. SLAM is run with DA3 keyframe depth, a frontend NMS of 2, a frontend threshold of 24, $\beta=0.65$, and 48 backend iterations. We verify the resulting trajectories using the time-reversal shuttle test described in Section~\ref{sec:problem_analysis}. Per-frame focal estimates are unstable on zoom-heavy sequences, so we fit a single focal length per clip, discarding extreme estimates outside $[0.3\times,\,3\times]$ the image width. We further smooth the trajectories with a One Euro filter ($\texttt{min\_cutoff}=0.001$, $\texttt{beta}=0.5$, $\texttt{d\_cutoff}=1.0$) to suppress high-frequency jitter.

\paragraph{Step 2. Human reconstruction in camera space.}
We use PromptHMR-Vid~\citep{wang2025prompthmr} as the default backbone, which models the camera with full perspective at an estimated per-clip focal length and supervises its camera-space translation against ground-truth translation. Its native world-space reconstruction follows TRAM, composing with DROID-SLAM~\citep{teed2021droid} and Metric3D~\citep{yin2023metric3d}, while GVHMR~\citep{shen2024world} builds its world trajectory by integrating a predicted root velocity, so for GVHMR in Table~\ref{tab:exp_backends} we supply the composition of Eq.~\ref{eq:composition} ourselves. We reconstruct humans in camera coordinates using a two-pass tracking-and-refinement procedure. We first select an initialization frame capturing the largest number of visible performers with minimal occlusion, then run SAM2.1~\citep{ravi2024sam2} to obtain initial tracks. In a second pass, we refine the tracks by prompting in both forward and backward directions using YOLO11x-pose~\citep{yolo11_ultralytics} bounding box detections. We then linearly interpolate bounding boxes across gaps of up to 5 frames, and the HMR model predicts meshes on the interpolated boxes.

\paragraph{Step 3. Camera-to-world composition.}
Given camera-space human predictions $\mathbf{X}$ and camera pose estimates $\{R_i, T_i\}$, the initial world-space reconstruction follows the standard transformation (Equation~\ref{eq:composition}). We apply that transformation to every predicted joint, though only the contact joints feed the scale recovery of Step~5. We then apply gravity correction, rotating the recovered world frame so its vertical axis follows the gravity direction given by the first-frame camera pitch and roll estimated by SPEC~\citep{SPEC:ICCV:2021}. We fit the floor height by RANSAC over the mesh vertices of all performers and place the floor at height zero, which fixes the floor plane that the grounding terms of Appendix~\ref{app:losses} measure against.

\paragraph{Step 4. Stationary-interval detection.}
The relative scale $s$ cannot be estimated while a performer is moving, since the performer's world-space motion $\Delta W_{\mathrm{obj}}$ in Eq.~\ref{eq:decomp} is then unknown and cannot be separated from the drift. The backbone reports translation at the pelvis, which keeps moving even while a performer holds position. A planted foot does not, so the contact joints $\mathcal{C}$ carry the only stillness we can read. We therefore look for intervals in which at least one foot stays planted, scoring each frame by the strongest contact confidence over $\mathcal{C}$ and admitting an interval once $\bar{c}_{p,t}$ clears $\tau_c$ on at least a fraction $\alpha = 0.8$ of its frames. Intervals are 10-frame windows slid one frame at a time. Two further gates remove intervals that pass the contact test yet still break the decomposition, one for displacement running across the camera direction rather than along it (Section~\ref{sec:method}), and one for rotation error large enough to compete with the translation (Appendix~\ref{app:rotation}). Inside an admitted interval we take $\Delta W_{\mathrm{obj},p,ij} \approx 0$, so the observed displacement is the drift $(1-s)\Delta T_{ij}$ of Eq.~\ref{eq:stationary_drift}. Broadcast cameras translate in all three dimensions rather than along one dominant axis, so we fit a single scale jointly over $x$, $y$ and $z$ instead of solving per axis and pooling afterwards.

\label{app:contact_input}
A backbone without a contact head would have to detect stillness from the composed trajectory, using foot speed or foot height above an estimated ground plane. This is circular. The composed trajectory already contains the drift we want to detect, and the drift grows with camera displacement, so a moving foot in that trajectory may reflect drift rather than performer motion. Contact confidence avoids this. It is predicted in camera space before composition adds the drift. Relying on contact also does not limit the choice of backbone. A backbone without a contact head can take contact confidence from another model. For TRAM, we take it from PromptHMR (Table~\ref{tab:exp_backends}).

\paragraph{Step 5. Closed-form scale recovery.}
\label{app:pipeline_step5}
Within the stationary intervals identified above, we estimate a single sequence-level scale $s^*$ by weighted least squares using Eq.~\ref{eq:closed_form_solution}. If no interval is admitted, we skip the correction and keep the uncorrected composition of Eq.~\ref{eq:composition}. Otherwise, before any optimization, we apply Eq.~\ref{eq:corrected_world}, measuring camera translation from the first frame, so the camera displacement $T_i - T_0$ is rescaled by $s^*$ and $(1-s^*)(T_i - T_0)$ is subtracted from every performer's world-space translation. The correction also shifts performers vertically, so we refit the floor by RANSAC over the mesh vertices of frames in which both feet are planted, and shift the performers and camera together to return the floor to height zero. We fill remaining translation gaps by linear interpolation in 3D world space and pose gaps by spherical linear interpolation. At both temporal boundaries where detections are absent, we extrapolate translation with the mean velocity of the last 10 observed frames, damped by a factor of 0.8 per frame, and hold the pose fixed. Contact logits more than 15 frames from any detection are blended toward a logit of 1, reaching it at 25 frames, so PGO treats these stretches as planted. For the HAC~\citep{yang2025humans} baseline, we replace $s^*$ with the mean of HAC's per-person scales, each estimated from PromptHMR-Vid joints against the ViPE point cloud, apply it through Eq.~\ref{eq:corrected_world} without PGO, and keep HAC's Gaussian smoothing of translation ($\sigma=3$ frames). On the paired reference, this lowers WA-MPJPE from 75.98 to 67.92\,cm, W-MPJPE from 160.43 to 121.38\,cm, and Pair Distance from 76.85 to 61.28\,cm relative to the uncorrected composition, and moves the ratio from 0.54 to 0.64.

\paragraph{Step 6. Physically grounded optimization.}
\label{app:pipeline_step6}
We refine the recovered world-space motion to enforce physical plausibility. As detailed in the following section, the objective combines reprojection consistency with translation smoothness, contact-aware grounding terms, and temporal inter-person repulsion, reducing foot skating, improving temporal coherence, and discouraging implausible multi-person collisions while preserving agreement with the input video.

\begin{table}
    \caption{\textbf{The K-Pop stress test is the only in-the-wild benchmark of broadcast camera work.}
    Camera motion is categorized as static (S), moving (M), and zooming (Z). AIST++~\citep{li2021ai} reports 10.1M images from 9 views, so this count is not directly comparable to monocular video frame totals. DCM~\citep{wang2024dancecamera3d} reports 193 minutes at 30 FPS, and DanceTrack~\citep{sun2022dancetrack} provides bounding boxes but not pose annotations.}
    \label{sup-tab:dataset_comparison}
    \centering
    \footnotesize
    \resizebox{\linewidth}{!}{%
    \begin{tabular}{l c c c c l c c c}
    \toprule
    \multirow{2}{*}{\textbf{Dataset}} & \multirow{2}{*}{\textbf{Setting}} & \multirow{2}{*}{\textbf{Domain}} & \multirow{2}{*}{\textbf{Camera motion}} & \multicolumn{2}{c}{\textbf{Annotation}} & \multicolumn{3}{c}{\textbf{Statistics}} \\
    \cmidrule(lr){5-6}
    \cmidrule(lr){7-9}
    & & & & \textbf{GT Pose} & \textbf{Source} & \textbf{People/Frame} & \textbf{Frames} & \textbf{Pose Instances} \\
    \midrule
    H3.6M~\citep{ionescu2013human3}       & Controlled  & Everyday     & S   & Yes & Calibrated     & 1        & 630k      & 630k \\
    3DPW~\citep{von2018recovering}       & In-the-wild & Everyday     & M   & Yes & IMU video      & 1--2     & 53k       & 75k \\
    EMDB~\citep{kaufmann2023emdb}        & In-the-wild & Everyday     & M   & Yes & EM sensors     & 1        & 105k      & 105k \\
    BEDLAM-2~\citep{tesch2025bedlam2}      & Synthetic   & Everyday     & M   & Yes & Synthetic      & $>1$     & 380k      & 1M \\
    AIST++~\citep{li2021ai}      & Controlled  & Dance        & S   & Yes & Calibrated     & $\sim$1  & 10.1M$^\ast$ & --- \\
    AIOZ-GDANCE~\citep{le2023music} & In-the-wild & Group dance  & --- & No  & ---            & $>1$     & 1.8M      & --- \\
    WorldPose~\citep{jiang2024worldpose}   & In-the-wild & Sports       & M+Z & Yes & LiDAR multicam & $>1$     & 150k      & 2.5M \\
    DanceTrack~\citep{sun2022dancetrack}  & In-the-wild & Casual       & M+Z & No  & Reconstructed  & $>1$     & 105k      & --- \\
    DCM~\citep{wang2024dancecamera3d}         & Edited      & Single dance & M+Z & Yes & Animator       & 1        & $\sim$347k & $\sim$347k \\
    \midrule
    \textbf{K-Pop stress test (ours)} & \textbf{In-the-wild} & \textbf{Broadcast} & \textbf{M+Z} & \textbf{No} & \textbf{Reconstructed} & \textbf{1--12} & \textbf{59.19k} & \textbf{336.87k} \\
    \bottomrule
    \end{tabular}%
    }
\end{table}

\section{Physically Grounded Losses}
\label{app:losses}

We expand the objective minimized in the physically grounded optimization (PGO) of Step~6.

\textbf{Indices.} \(p\) indexes performers, \(t = 0, \ldots, T-1\) frames, \(j\) body joints, \(a\) and \(b\) the body joints approximated by collision spheres, and \(d \in \{x,y\}\) the two image axes. \(\mathcal{J}\) is the set of body joints scored for reprojection, the 25 joints of the OpenPose layout without the mid-hip and the two hips. \(\mathcal{C}\) is the set of contact joints, the left and right ankles and feet. \(\mathcal{B}\) is the set of all 29 body joints, the 25 OpenPose joints together with \(\mathcal{C}\).

\textbf{Quantities.} \(Y_{p,t,j} \in \mathbb{R}^3\) are the optimized world-space joints and \(x_{p,t} \in \mathbb{R}^3\) the optimized world-space translation offset. \(\hat{u}_{p,t,j} \in \mathbb{R}^2\) are the 2D keypoints detected by ViTPose~\citep{xu2022vitpose} and \(q_{p,t,j}\) their detection confidence. \(m_{p,t} \in \{0,1\}\) marks the frames on which performer \(p\) has a track. Step~5 fills every gap and extrapolates to both clip boundaries, so \(m_{p,t} = 1\) throughout, and frames without a detection drop out of \(\mathcal{L}_{\mathrm{kp}}\) through their zero detection confidence. \(h_{p,t}\) is the bounding-box height and \(f = 30\) a fixed time-scaling constant. The map \(\pi(\cdot)\) projects a world-space point into the image, and \(\pi_d(\cdot)\) returns its \(d\) component. Contact confidences are \(c_{p,t,k} = \sigma(\ell_{p,t,k})\) for backbone logits \(\ell_{p,t,k}\) and sigmoid \(\sigma(\cdot)\), as in Section~\ref{sec:method}.

The PGO objective is
{\small\[
\mathcal{L}
=
\lambda_{\mathrm{kp}} \mathcal{L}_{\mathrm{kp}}
+
\lambda_{\mathrm{sm}} \mathcal{L}_{\mathrm{sm}}
+
\lambda_{\mathrm{rep}} \mathcal{L}_{\mathrm{rep}}
+
\lambda_{\mathrm{floor}} \mathcal{L}_{\mathrm{floor}}
+
\lambda_{\mathrm{cv}} \mathcal{L}_{\mathrm{cv}}
+
\lambda_{\mathrm{ch}} \mathcal{L}_{\mathrm{ch}}
+
\lambda_{\mathrm{sf}} \mathcal{L}_{\mathrm{sf}},
\]}
where each \(\lambda\) is a scalar weight on its term, with \(\lambda_{\mathrm{kp}} = 1.5\),
\(\lambda_{\mathrm{sm}} = 1\), \(\lambda_{\mathrm{rep}} = 10\), \(\lambda_{\mathrm{floor}} = 100\),
\(\lambda_{\mathrm{cv}} = 500\), \(\lambda_{\mathrm{ch}} = 10\), and \(\lambda_{\mathrm{sf}} = 100\). \(\mathcal{L}_{\mathrm{cv}}\), \(\mathcal{L}_{\mathrm{ch}}\), and \(\mathcal{L}_{\mathrm{sf}}\) are contact-aware terms, enabled only when contact predictions are available. We minimize \(\mathcal{L}\) with Adam for 3000 iterations, with learning rate \(10^{-2}\) for \(x_{p,t}\) and for a camera rotation and vertical camera offset shared across frames, and \(5 \times 10^{-2}\) for the residual scale \(\Delta s\) and floor offset \(b\) introduced below.

\textbf{Reprojection term.}
To preserve image consistency during world-space refinement, we penalize the discrepancy between
projected 3D joints and detected 2D keypoints.
Let \(\Delta u_{p,t,j,d} = \pi_d(Y_{p,t,j}) - \hat{u}_{p,t,j,d}\) denote the per-axis reprojection residual.
We combine a Geman--McClure robust penalty \(\rho(\cdot)\) with an L1 term
and normalize by bounding-box height so that scale is comparable across performers and frames:
{\small\[
\mathcal{L}_{\mathrm{kp}}
=
\frac{1}{\sum_{p,t} m_{p,t}}
\sum_{p,t} m_{p,t}
\cdot
\frac{1}{|\mathcal{J}|}
\sum_{j \in \mathcal{J}}
\mathbf{1}[q_{p,t,j} > 0.9]
\cdot
\frac{1}{2h_{p,t}}
\sum_{d \in \{x,y\}}
\Bigl[
  \rho(\Delta u_{p,t,j,d}) + |\Delta u_{p,t,j,d}|
\Bigr],
\]}
where \(\rho(u) = \kappa^2 u^2 / (\kappa^2 + u^2)\) with \(\kappa = 150\) pixels, applied before the division by \(h_{p,t}\).
Only keypoints detected with confidence above \(0.9\) contribute.

\textbf{Translation smoothness.}
We regularize the optimized world-space translation to discourage temporally inconsistent motion. The closed-form estimate \(s^{*}\) of Eq.~\ref{eq:closed_form_solution} is approximate, so PGO learns a single residual scale \(\Delta s\) on top of it, giving an effective relative scale \(s^{*} + \Delta s\). The final per-performer translation is
{\small\[
\tilde{x}_{p,t} = x_{p,t}^{(0)} + x_{p,t} + \Delta s\,(T_t - T_0) - b\,e_y,
\]}
where \(x_{p,t}^{(0)}\) is the initial translation after the closed-form correction of Step~5, \(x_{p,t}\) the learned per-frame offset, \(T_t\) the camera translation at frame \(t\), \(b\) a learned floor offset, and \(e_y\) the vertical unit vector. The camera receives the same shift \(\Delta s\,(T_t - T_0) - b\,e_y\), so \(\Delta s\) and \(b\) move the camera and all performers together and leave every reprojection unchanged. They are set by the smoothness, contact, and floor terms. We penalize both scaled velocity (squared) and scaled acceleration (unsquared norm):
{\small\[
\mathcal{L}_{\mathrm{sm}}
=
\frac{
  \sum_{p,t=1}^{T-1} m_{p,t}\,\dfrac{f^2}{3}\|\Delta\tilde{x}_{p,t}\|_2^2
}{
  \sum_{p,t=1}^{T-1} m_{p,t}
}
+
\frac{
  \sum_{p,t=1}^{T-2} m_{p,t}\,f\,\|\Delta^2\tilde{x}_{p,t}\|_2
}{
  \sum_{p,t=1}^{T-2} m_{p,t}
},
\]}
where \(\Delta\tilde{x}_{p,t} = \tilde{x}_{p,t} - \tilde{x}_{p,t-1}\),
\(\Delta^2\tilde{x}_{p,t} = \tilde{x}_{p,t+1} - 2\tilde{x}_{p,t} + \tilde{x}_{p,t-1}\),
and the factor of \(f^2/3\) arises from time-scaling followed by
mean-reduction over the three spatial coordinates.

\textbf{Inter-person repulsion.}
To reduce implausible interpenetration in crowded scenes, we approximate 12 body joints by
spheres with centers \(v_{p,t,a}\) and radii \(r_a\), namely the neck, shoulders, hips, and knees
(\(r_a = 0.05\)\,m), the mid-hip (\(0.06\)\,m), the elbows (\(0.035\)\,m), and the wrists (\(0.03\)\,m).
For each pair of performers \(p < q\), the framewise penetration energy is
{\small\[
e_{pq,t}
=
\sum_{a,b}
\left[
  r_a + r_b + \delta - \|v_{p,t,a} - v_{q,t,b}\|_2
\right]_+^2,
\]}
where \(\delta = 0.2\)\,m is a safety margin.
We define a soft gate \(g_{pq,t} = e_{pq,t}/(e_{pq,t} + \varepsilon)\), with \(\varepsilon = 10^{-4}\), that focuses the temporal penalty on frames with active overlap, and a
triangular window \(\omega_\tau = \Omega + 1 - \tau\) over offsets \(\tau = 1,\ldots,\Omega\), with \(\Omega = 5\).
The combined loss is
{\small\[
\mathcal{L}_{\mathrm{rep}}
=
\mu\,
\frac{\sum_{p<q,t}\mathbf{1}[e_{pq,t}>0]\,e_{pq,t}}
     {\sum_{p<q,t}\mathbf{1}[e_{pq,t}>0]}
+
(1-\mu)\,
\frac{\sum_{\tau=1}^{\Omega} \omega_\tau\,
      \dfrac{\sum_{p<q,t}\,g_{pq,t-\tau}(e_{pq,t}-e_{pq,t-\tau})^2}
            {\sum_{p<q,t}\,g_{pq,t-\tau}}
     }{\sum_{\tau=1}^{\Omega} \omega_\tau},
\]}
which combines a direct overlap penalty, averaged only over penetrating frames, with a
temporally-gated term that discourages abrupt frame-to-frame changes in collision energy.
The scalar \(\mu = 0.5\) balances the two.

\textbf{Contact-aware motion losses.}
When a joint is predicted to be in contact, its world-space motion should remain small over time.
We therefore use a soft contact-velocity penalty
{\small\[
\mathcal{L}_{\mathrm{cv}}
=
\frac{1}{|\mathcal{C}|}
\cdot
\frac{
  \sum_{p,t}\, m_{p,t+1}
  \sum_{k \in \mathcal{C}} c_{p,t+1,k}\,\|Y_{p,t+1,k} - Y_{p,t,k}\|_2^2
}{
  \sum_{p,t} m_{p,t+1}
},
\]}
which suppresses motion proportionally to the predicted contact confidence and normalizes jointly
by the number of contact joints and the count of tracked performer-frames.
To further reduce foot skating, we also impose a stronger high-confidence stickiness term
{\small\[
\mathcal{L}_{\mathrm{sf}}
=
\frac{
  \sum_{p,t,k \in \mathcal{C}}
  m_{p,t+1}\,\mathbf{1}[c_{p,t+1,k} > \tau_c]\,\|Y_{p,t+1,k} - Y_{p,t,k}\|_1
}{
  \sum_{p,t,k \in \mathcal{C}}
  m_{p,t+1}\,\mathbf{1}[c_{p,t+1,k} > \tau_c]
},
\]}
where \(\tau_c = 0.5\) is the contact threshold of Section~\ref{sec:method}.
The denominator counts only the active high-confidence contact detections, providing a
self-normalizing objective robust to variation in the number of contact events.

\textbf{Grounding terms.}
We encourage predicted contact joints to remain close to the ground plane:
{\small\[
\mathcal{L}_{\mathrm{ch}}
=
\frac{1}{|\mathcal{C}|}
\cdot
\frac{
  \sum_{p,t}\, m_{p,t}
  \sum_{k \in \mathcal{C}} c_{p,t,k}\,\bigl|Y_{p,t,k}^{(y)} - y_{\mathrm{c}}\bigr|
}{
  \sum_{p,t} m_{p,t}
},
\]}
where \(Y_{p,t,k}^{(y)}\) denotes the vertical coordinate. The floor lies at height zero, and
\(y_{\mathrm{c}} = 0.08\)\,m is the target height of a contact joint above it.
To prevent anatomically implausible solutions, we penalize any joint that moves below the floor:
{\small\[
\mathcal{L}_{\mathrm{floor}}
=
\frac{1}{|\mathcal{B}|}
\cdot
\frac{
  \sum_{p,t} m_{p,t} \sum_{j \in \mathcal{B}} \bigl[-Y_{p,t,j}^{(y)}\bigr]_+
}{
  \sum_{p,t} m_{p,t}
}.
\]}

Together, these terms balance image consistency, temporal smoothness, collision avoidance,
contact-consistent motion, and floor feasibility.

\section{Drift Assumptions and Error Sources}
\label{app:drift_assumption}
The closed-form scale estimator in Eq.~\ref{eq:closed_form_solution} assumes that the world-space displacement of a performer within a stationary interval comes entirely from the scale mismatch, $(1-s)\Delta T_{ij}$ in Eq.~\ref{eq:stationary_drift}. Two conditions have to hold for that. (1) The performer must be genuinely stationary, which the contact gate of Step~4 in Appendix~\ref{app:pipeline} enforces, and (2) error in the estimated camera rotation must be small relative to the camera translation, which Appendix~\ref{app:rotation} enforces. Appendix~\ref{app:gates} lists the remaining interval gates. Appendix~\ref{app:other_sources} examines which upstream errors feed the bias the estimator absorbs and which do not explain the observed drift.

\subsection{Rotation Error}
\label{app:rotation}
Eq.~\ref{eq:decomp} compares the two camera-to-performer vectors after rotating each into world axes (Section~\ref{sec:problem_analysis}), so camera rotation affects the scale estimate through the error in its estimate. If the estimated rotation at one end of an interval is off by a small angle, the camera-to-performer vector of a performer at camera-space depth $d$ shifts by roughly that angle times $d$, and Eq.~\ref{eq:stationary_drift} reads the shift as drift. When this shift is comparable to the translational signal $\|\Delta T\|$, the scale estimate becomes biased. We therefore compute a per-interval error budget and exclude intervals in which the expected rotational error exceeds the translational signal by a defined margin.

\paragraph{Relative rotation and angular error.}
For each candidate interval $[t_0,\,t_1]$, we compute the relative camera rotation
{\small
\[
R_{\mathrm{rel}} = R_{t_1}\,R_{t_0}^{\top},
\]
}
where $R_t$ is the camera-to-world rotation at frame $t$ of Section~\ref{sec:background}, and extract the geodesic angle in radians via
{\small
\[
\theta = \arccos\!\left(\frac{\mathrm{tr}(R_{\mathrm{rel}}) - 1}{2}\right).
\]
}We model the rotation error of visual odometry as a fixed noise floor plus a fraction of the rotation it must recover,
{\small
\[
\theta_{\mathrm{err}} = \theta_0 + 0.1\,\theta,
\]
}where $\theta_0 = 0.5^\circ$ is the noise floor of the rotation estimator. All angles are in radians, so $\theta_{\mathrm{err}}\,d$ in the criterion below is a displacement.

\paragraph{Interval exclusion criterion.}
We exclude an interval when the displacement expected from rotation error at the performer's camera-space depth $d$, scaled by a margin $\eta$, exceeds the camera translation magnitude $\|\Delta T\|$,
{\small
\[
\theta_{\mathrm{err}}\cdot d \cdot \eta \;>\; \|\Delta T\|.
\]
}Equivalently, we require the signal-to-noise ratio $\|\Delta T\| / (\theta_{\mathrm{err}} d)$ to reach at least $\eta$. We set $\eta = 1.5$, so the translational signal must exceed the expected rotational error by that factor before an interval contributes to the scale estimate.

\subsection{Additional Interval Gates}
\label{app:gates}
We admit an interval when at least a fraction $\alpha = 0.8$ of its frames pass the contact threshold $\tau_c$. We discard intervals in which the camera translates less than 1\,mm, since they may be noise, and samples from performers closer than 0.1\,m to the camera, where depth is unreliable. Within the fit, we downweight axes along which the camera barely moved, for the same reason.

\subsection{Analysis of Drift Sources}
\label{app:other_sources}
\begin{figure}[!h]
    \centering
    \includegraphics[width=0.95\linewidth]{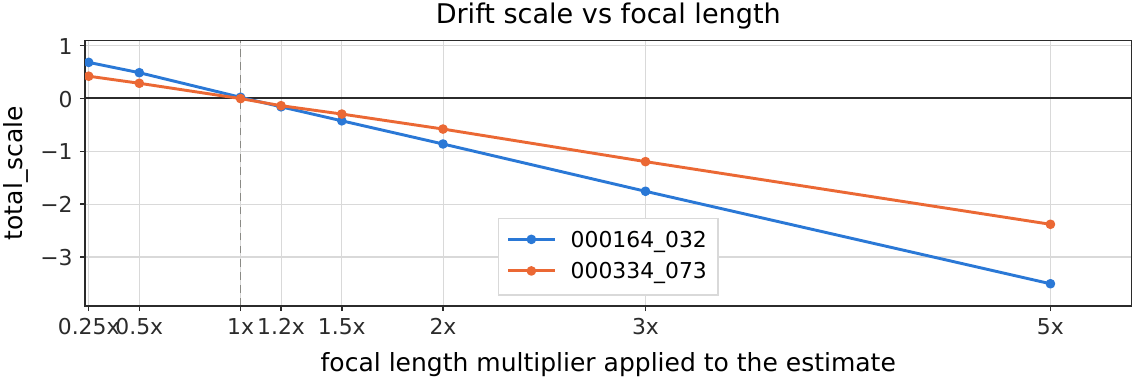}
    \caption{\textbf{Drift scale under a rescaled focal length.} We multiply the backend-estimated focal length by a fixed multiplier and rerun the correction. Each curve is one clip and shows the change in estimated drift fraction relative to the unperturbed focal length. The change is linear in the multiplier, as expected when focal error acts as a scale degree of freedom, and its slope differs between clips because each carries a different scale.}
    \label{fig:focal_ablation}
\end{figure}

With body scale fixed by the body model, a focal length too large by a factor recovers a depth too large by the same factor. Scaling all depths together with the camera translation leaves image projections unchanged. Focal error and depth ambiguity are therefore scale degrees of freedom that hold across a clip, and Eq.~\ref{eq:closed_form_solution} estimates the resulting bias, whichever of them produced it. Figure~\ref{fig:focal_ablation} shows that focal length error acts as this scale degree of freedom, since an incorrect focal length shifts the estimated bias in proportion to the error, and the correction removes that shift.

Crop-induced translation instability and detector-induced scale variation act on the camera-space estimate with a magnitude and direction that change between frames. PromptHMR-Vid~\citep{wang2025prompthmr} operates without crops while GVHMR~\citep{shen2024world} uses them, yet both drift and both respond to the same correction (Table~\ref{tab:exp_backends}), so we treat these effects as residual error and not as an account of the drift. Where such an error carries a component that holds across a sequence, it acts as a scale bias, and the correction absorbs it with the rest.
\section{Scale Estimation Ablations}
\label{app:scale_ablations}
The closed-form estimator in Eq.~\ref{eq:closed_form_solution} rests on two choices, which intervals are pooled and how many scales are fit from them. Table~\ref{tab:threshold} sweeps the contact threshold that admits an interval. Table~\ref{tab:scalemodel} replaces the single sequence-level scale with a time-varying one. 

\subsection{Sensitivity to the number of stationary intervals}
\label{app:window_sparsity}
Stationary moments can be scarce in choreography, so the sequence-level estimate may have few intervals to pool. On one sequence, we run the pipeline twice, once with all 125 admitted intervals and once with 3 of them, and record both the closed-form estimate and the final result (Table~\ref{tab:sparsity}).

\begin{table}[h]
    \centering
    \caption{\textbf{Cutting the pooled intervals from 125 to 3 changes the closed-form estimate but not the final result.} Both rows are the same sequence, and values are drift fractions $a = 1 - s$. $a^{*}$ is the closed-form estimate of Eq.~\ref{eq:closed_form_solution}, $\Delta a$ is the shift PGO applies through its learned residual scale, and $a$ is the value after both stages.}
    \label{tab:sparsity}
    \footnotesize
    \begin{tabular}{cccc}
    \toprule
    \textbf{Admitted intervals} & \textbf{$a^{*}$ (closed form)} & \textbf{$\Delta a$ (PGO)} & \textbf{$a$ (final)} \\
    \midrule
    125 & 0.310 & $+0.008$ & 0.318 \\
    3   & 0.443 & $-0.123$ & 0.320 \\
    \bottomrule
    \end{tabular}
\end{table}

The closed-form estimate moves by 0.133 between the two runs. PGO absorbs almost all of it, and the final values sit 0.002 apart. On this sequence, the number of pooled intervals does not determine the final result as the second stage recovers what the first misses.

The time-varying scale in Table~\ref{tab:scalemodel} therefore does not fail for lack of intervals. Each segment overfits the intervals it does have, the segment scales stop agreeing, and the reassembled camera path disturbs motion the visual odometry never observed. Since we place performers relative to that path, the spurious motion reaches their trajectories, which is why we fit one scale per sequence.

\begin{table}[h]
    \centering
    \caption{\textbf{Constant against time-varying scale.} A time-varying scale is the more expressive choice, yet it destabilizes the recovered world-space motion, and the damage falls on the temporal metrics while the 2D metrics hold. Bold marks the best value in each column. Both rows use ViPE~\citep{huang2025vipe} and PromptHMR-Vid~\citep{wang2025prompthmr}.}
    \label{tab:scalemodel}
    \begin{tabular}{lcccc}
    \toprule
    \multirow{2}{*}{\textbf{Scale model}} & \multicolumn{2}{c}{\textbf{2D Reprojection}} & \multicolumn{2}{c}{\textbf{3D World Space (mm)}} \\
    \cmidrule(lr){2-3}
    \cmidrule(lr){4-5}
    & \textbf{PCK} $\uparrow$ & \textbf{Mask IoU} $\uparrow$ & \textbf{Jitter} $\downarrow$ & \textbf{Foot Skating} $\downarrow$ \\
    \midrule
    Constant scale     & \textbf{0.868} & 0.720 & \textbf{26.87} & \textbf{17.94} \\
    Time-varying scale & 0.862 & \textbf{0.737} & 131.57 & 32.58 \\
    \bottomrule
    \end{tabular}
\end{table}

The same two-stage behavior explains Table~\ref{tab:threshold}. The contact threshold decides which intervals are pooled, so changing it changes the closed-form estimate, yet the reported metrics barely move. In both experiments, the first stage takes a large perturbation, and the learned residual absorbs what survives it. The stability in Table~\ref{tab:threshold} is a property of the two stages together.

\begin{table}[h]
    \centering
    \caption{\textbf{The reconstruction is robust to the contact threshold $\tau_c$.} The threshold sets which intervals enter the estimate. Across the tested range, with ViPE~\citep{huang2025vipe} and PromptHMR-Vid~\citep{wang2025prompthmr}, the 2D metrics and jitter barely move, and foot skating rises modestly. Every setting stays best on foot skating and second on jitter among the methods of Table~\ref{tab:exp_comparison}, behind only HAC~\citep{yang2025humans}. We use $\tau_c = 0.5$.}
    \label{tab:threshold}
    \begin{tabular}{lcccc}
    \toprule
    \multirow{2}{*}{\textbf{Contact threshold} $\tau_c$} & \multicolumn{2}{c}{\textbf{2D Reprojection}} & \multicolumn{2}{c}{\textbf{3D World Space (mm)}} \\
    \cmidrule(lr){2-3}
    \cmidrule(lr){4-5}
    & \textbf{PCK} $\uparrow$ & \textbf{Mask IoU} $\uparrow$ & \textbf{Jitter} $\downarrow$ & \textbf{Foot Skating} $\downarrow$ \\
    \midrule
    0.3           & 0.866 & 0.718 & 26.34 & 16.83 \\
    0.5 (default) & 0.868 & 0.720 & 26.87 & 17.94 \\
    0.8           & 0.872 & 0.726 & 26.38 & 19.09 \\
    \bottomrule
    \end{tabular}
\end{table}

\subsection{Stationary-performer clips}
\label{app:stationary}
This ablation isolates the correction on clips where the expected answer is known, using 10 clips in which 60 performers hold their stage positions, so the true world-space displacement is near zero and whatever motion is recovered is error. Table~\ref{tab:stationary} separates what each stage contributes. PGO removes most of the raw drift on its own, cutting net displacement from 2.334~m to 0.581~m, and adding the drift correction reduces it to 0.153~m. Jitter barely moves between those two rows, so the further reduction comes from placing the performers correctly rather than from just smoothing their trajectories.

\begin{table}[h]
    \centering
    \caption{\textbf{Drift correction reduces displacement.} On 10 clips covering 60 performers who hold their stage positions, the expected world-space displacement is near zero, so smaller is closer to the truth. Jitter is unchanged between the last two rows, which is why it carries no bold entry. Bold marks the best value in each column. All rows use ViPE~\citep{huang2025vipe} and PromptHMR-Vid~\citep{wang2025prompthmr}.}
    \label{tab:stationary}
    \footnotesize
    \begin{tabular}{lcccc}
    \toprule
    \multirow{2}{*}{\textbf{Configuration}} & \multicolumn{2}{c}{\textbf{World-Space Displacement (m)}} & \multicolumn{2}{c}{\textbf{Stability (mm)}} \\
    \cmidrule(lr){2-3} \cmidrule(lr){4-5}
    & \textbf{Path length} $\downarrow$ & \textbf{Net displacement} $\downarrow$ & \textbf{Jitter} $\downarrow$ & \textbf{Foot Skating} $\downarrow$ \\
    \midrule
    Raw trajectory        & 5.206 & 2.334 & 105.128 & 47.269 \\
    PGO                   & 0.849 & 0.581 & 12.465 & 7.568 \\
    PGO + drift correction & \textbf{0.461} & \textbf{0.153} & \textbf{12.379} & \textbf{4.312} \\
    \bottomrule
    \end{tabular}
\end{table}

\section{K-Pop Stress Test Construction Details}
\label{app:dataset_construction}
We collect approximately 1,000 K-Pop broadcast dance videos from YouTube and standardize all video to 24 FPS at $1280 \times 720$ resolution. We detect hard cuts with TransNetV2~\citep{soucek2024transnet} and remove four frames on each side of every transition to improve temporal consistency. Broadcast videos frequently alternate between close-up shots and wide stage views, so we filter for clips with stable multi-person coverage suitable for world-space evaluation. Using YOLO11n-Pose detections, we discard frames dominated by upper-body close-ups (detections covering more than 25\% of the image height), clips with at least 0.5 seconds of consecutive missed detections, clips containing only a single visible performer, and clips shorter than 3 seconds. We then pass surviving clips through the time-reversal shuttle test described in Section~\ref{sec:problem_analysis} to validate camera trajectory quality.

\paragraph{Data availability and ethical considerations.}
We do not redistribute raw video. Upon release, we will provide the YouTube URLs and timestamps required to reproduce the dataset, with downloading left to the end user in accordance with the platform's terms of service.

\section{User Study Design}
\label{app:user_study}
\begin{figure}[h]
    \centering
    \includegraphics[width=0.6\linewidth]{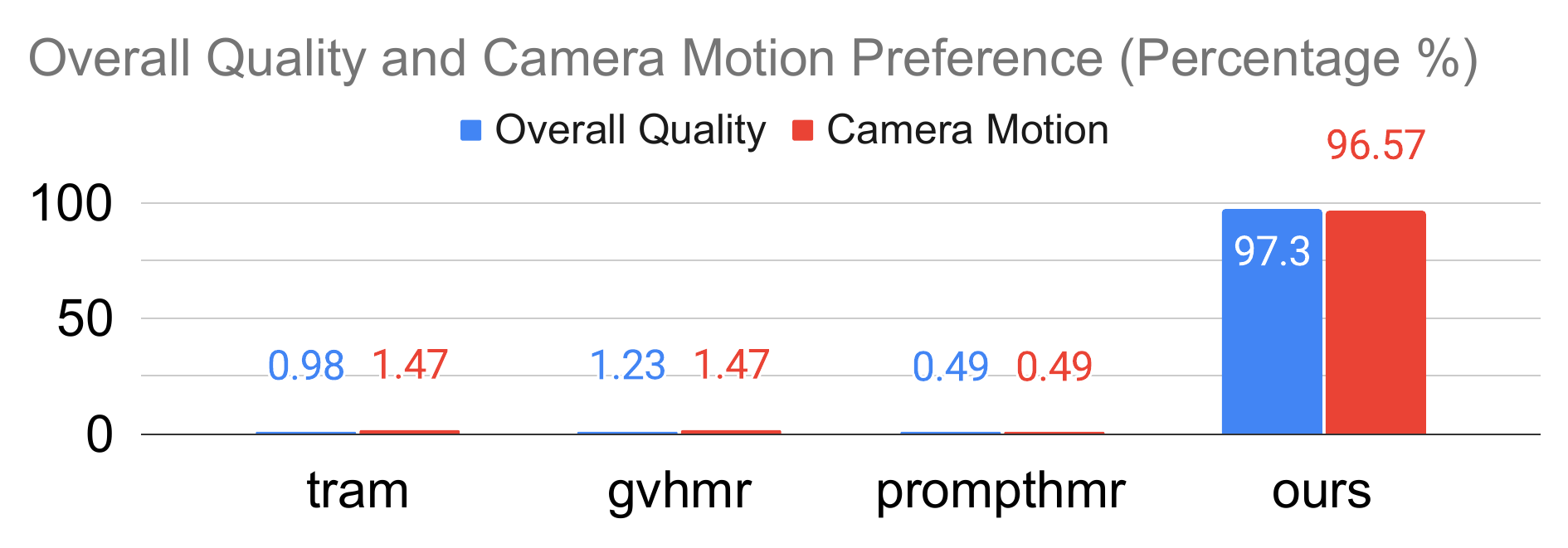}
    \caption{\textbf{User study.} Share (\%) of all votes each method received on overall quality and camera motion, pooled over 24 pairwise comparisons from 34 participants. Each baseline (TRAM~\citep{wang2024tram}, GVHMR~\citep{shen2024world}, and PromptHMR~\citep{wang2025prompthmr}) appears in one third of the comparisons.}
    \label{fig:user_study}
\end{figure}

We conducted a perceptual study across 8 sequences with 34 participants. Each comparison video showed five synchronized panels, the original input, a 2D reprojection of colored meshes overlaid on the input, a front-facing 3D view, a top-down 3D view, and a trajectory plot showing each performer's path and the estimated camera position. Participants answered two questions per comparison, which result provides more consistent overall quality across 2D alignment and 3D realism, and which better matches the camera motion in the input video. The question order was randomized. Since our method was compared against three baselines independently, each sequence appeared three times throughout the survey, yielding 24 comparisons per participant and 816 votes per question. As shown in Figure~\ref{fig:user_study}, our method received 97.3\% and 96.6\% of all votes on overall quality and camera motion, respectively. Against each baseline individually, it was preferred in at least 96.3\% of comparisons on overall quality and 95.6\% on camera motion.

\end{document}